\documentclass[a4paper,fleqn]{cas-dc}

\usepackage[authoryear]{natbib}
\usepackage{tabularx}  
\usepackage{booktabs}
\usepackage{siunitx}  
\usepackage{placeins}
\usepackage{amsmath} 
\usepackage{array}  
\usepackage{soul,color}

\makeatletter
\def\ps@first{%
  \let\@oddhead\@empty
  \let\@evenhead\@empty
  \let\@oddfoot\@empty
  \let\@evenfoot\@empty
}
\makeatother

\makeatletter
\def\printorcid{}
\makeatother

\begin{document}
\let\WriteBookmarks\relax
\def\floatpagepagefraction{1}
\def\textpagefraction{.001}

\shorttitle{Towards Graph-Based Deep Learning for Map Generalization}

\shortauthors{Y. Wang et al.}

\title [mode = title]{Towards Graph-Based Deep Learning for Map Generalization: Insights from Building Footprints Simplification and Aggregation}



%

\author[1]{Yanning Wang}
\cormark[1]
\ead{ywang979@connect.hkust-gz.edu.cn}
\cortext[cor1]{Corresponding author}

\author[2]{Zhiyong Zhou}
\ead{zhiyong.zhou@geo.uzh.ch}

\author[1]{Zhouyu Liu}
\ead{zliu382@connect.hkust-gz.edu.cn}

\author[1]{Mengni Yu}
\ead{myu748@connect.hkust-gz.edu.cn}

\author[3]{Yu Feng}
\ead{y.feng@tum.de}

\affiliation[1]{organization={The Hong Kong University of Science and Technology (Guangzhou)},
    country={China}}
    
\affiliation[2]{organization={Zhejiang University},
    country={China}}

\affiliation[3]{organization={Mainz University of Applied Sciences},
    country={Germany}}
    








\begin{abstract}
Map generalization remains one of the fundamental tasks in cartography, especially for the simplification and aggregation of complex building footprints.
This study presents the first exploratory application of graph-based deep learning to both tasks, reformulating simplification as node movement prediction and aggregation as link prediction within a unified graph learning framework.
We evaluate representative graph neural network architectures (GCN, GAT, and GraphSAGE) on multi-scale building datasets, showing that GraphSAGE demonstrates relative strengths in link prediction accuracy, while also revealing persistent challenges in precise node movement prediction.
Beyond quantitative performance, the results highlight that aggregation poses greater complexity and challenges than simplification, underscoring the difficulty of capturing higher-level spatial relationships in map generalization with current deep learning approaches.
Although limitations such as data imbalance and the need for post-processing remain, the study provides valuable insights and methodological directions for advancing automated map generalization with deep learning approaches.

\end{abstract}



\begin{keywords}
Map generalization \sep Simplification and aggregation  \sep Graph deep learning models \sep Building footprints \sep Node movement \sep Link prediction 
\end{keywords}
\maketitle
\pagestyle{empty}     

\section{Introduction}
Map generalization is one of the most fundamental tasks in cartography. As an essential process in spatial data handling and map production, map generalization abstracts geographic information to facilitate representation at reduced scales while preserving central spatial relationships and characteristics, and also reduces data volume \citep{weibel1999generalising}. Ideally, the generalized output should provide a visually comprehensible rendition of complex geographic data. The automation of map generalization is highly desired to reduce the manual efforts of expert cartographers and improve the efficiency of map production \citep{Brassel1988, Mackaness2007}. The classical approaches for automated map generalization are mainly algorithmic with certain procedures and parameters. Despite decades of research, however, automated map generalization remains challenging because the complex spatial distribution and diverse visual perception of map objects, such as buildings and roads, make it difficult to generalize maps with one-size-fits-all parametric algorithms \citep{mackaness2014automation}. 

Recent advances in deep learning provide promising opportunities for map generalization, with approaches proposed using rasterized vector inputs \citep{sester2019deep, touya2019deep} or direct vector inputs \citep{Zhou2023}. Among these, vector-based approaches show greater potential due to their explicit representation of geometries, which avoids the accuracy loss inherent in rasterization.

For the vector-based approaches, graph representations are particularly suitable, as they can capture spatial context by capturing relationships between map objects—such as proximity, relevance, and structural importance \citep{Mackaness1993}. 
With such vector representations, Graph Neural Networks (GNNs), and in particular Graph Convolutional Networks (GCNs) \citep{kipf2017gcn}, have been applied to tasks such as building shape encoding and footprint simplification in map generalization. 
In particular, \citet{Zhou2023} demonstrated that GCNs can assist in simplification by learning to adjust or remove vertices in building polygons.
However, map generalization often requires considering the spatial relationships among multiple neighboring polygons or geo-entities, and thus, focusing solely on encoding and simplifying individual polygons is insufficient.

Building aggregation, despite being a common operator in map generalization, has received little attention in learning-based frameworks. Compared to simplification, aggregation is inherently more complex because it involves capturing higher-order relationships between adjacent buildings, reasoning about spatial proximity, and producing coherent merged structures \citep{shen2019polygon, zhang2023polygonal}. This complexity makes aggregation a particularly challenging task for graph-based learning, and its feasibility remains largely unexplored. Although some research has been dedicated to building group recognition, which is a preliminary step of building aggregation, with graph-based learning models \citep{yan2022graph}, studies on geometric operations for building aggregation are scarce.

The present work investigates the potential of a graph-based deep learning framework for map generalization, with a specific focus on building footprint simplification and aggregation. 
We propose a unified graph representation in which simplification is reformulated as a node movement task and aggregation as a link prediction task. To examine the feasibility of this approach, we implement and compare three representative GNN architectures—GCN, GAT, and GraphSAGE—on multi-scale building datasets, and analyze their performance with particular attention to the greater complexity of aggregation compared with simplification.


\section{Related Work}
\subsection{Simplification-focused Studies}
Simplification is one of the most fundamental operators in map generalization, aiming to reduce the geometric complexity of spatial objects while retaining their essential shapes and spatial patterns. Early research on simplification mainly focused on linear features. Classical algorithms such as Douglas–Peucker \citep{Douglas1973} and Visvalingam–Whyatt \citep{Visvalingam2017} reduced the number of vertices in polylines by iteratively eliminating redundant points according to geometric thresholds or effective areas. These methods provided efficient geometric solutions but often ignored cartographic requirements such as semantic preservation and topological consistency. Later studies attempted to address these limitations. For example, \citet{Wang1998} introduced bend-based operators to preserve structural properties by considering size, shape, and context, while \citet{Saalfeld1999} extended the Douglas–Peucker algorithm to handle topological conflicts. Such approaches improved the cartographic plausibility of simplified results but still relied on rigid rules and handcrafted parameters.

Beyond rule-based methods, optimization techniques were proposed to integrate multiple constraints into the simplification process. Approaches using gradient descent \citep{Ware1998}, simulated annealing \citep{Ware2003}, and genetic algorithms \citep{Wilson2003} enabled the search for optimal solutions that satisfy competing cartographic rules. Although these optimization-based models offered greater flexibility, they were computationally demanding and often lacked generalizability across diverse data types.

With the emergence of data-driven approaches, simplification has been increasingly studied from a machine learning perspective. For instance, supervised learning methods have been used to classify buildings or roads into categories such as retained, eliminated, or simplified \citep{Lee2017, duchene2018automatic}, thereby automating the operator selection process. More recently, deep learning techniques such as convolutional neural networks (CNNs) and graph convolutional networks (GCNs) have been applied to capture geometric structures and contextual relationships in vector data \citep{Ai2022, Zhou2023}. These models can learn implicit simplification rules directly from examples, improving the automation and adaptability of simplification tasks. However, most existing data-driven studies focus on either line or building simplification in isolation, and challenges remain in developing end-to-end frameworks that balance geometric reduction with semantic and topological preservation.

\subsection{Aggregation-focused Studies}
Aggregation is another essential operator in map generalization, which combines multiple nearby objects into a single, generalized unit while retaining the overall spatial pattern and functional semantics. Traditional aggregation methods typically rely on geometric and topological rules. For example, \citet{Guo2000} proposed interactive techniques to determine which buildings should be merged, with the algorithm subsequently carrying out the simplification and merging. Common approaches include merging topologically adjacent polygons or visually proximate objects, often guided by distance thresholds, shape similarity, or adjacency relationships. While effective in simple cases, these rule-based methods often fail to account for higher-level semantic consistency or complex urban morphologies.

To overcome such limitations, constraint-based and optimization models were developed. These approaches consider multiple cartographic requirements simultaneously, such as preserving settlement structure, functional connectivity, or density distribution \citep{Weibel1998}. By framing aggregation as a multi-objective optimization problem, algorithms such as simulated annealing and agent-based modeling \citep{Barrault2001} were introduced to handle trade-offs between geometric accuracy and cartographic legibility. Although more sophisticated, these methods remain sensitive to parameter choices and struggle to generalize across heterogeneous datasets.

More recently, data-driven approaches have emerged for aggregation. Machine learning techniques have been applied to classify groups of buildings for elimination, retention, or merging \citep{Lee2017, duchene2018automatic}. Graph-based models have proven particularly suitable, as they can encode spatial relationships among polygons. For example, Graph Convolutional Networks (GCNs) have been used to cluster buildings based on spatial arrangements and contextual attributes \citep{Zhou2023, Mai2023}. Such approaches highlight the potential of learning-based frameworks to capture both geometric and semantic cues in aggregation tasks. However, existing research tends to treat aggregation independently from other operators, and there is still a lack of integrated models that can jointly address simplification and aggregation while ensuring semantic coherence and topological consistency.

\subsection{Towards Integrated Frameworks with Graph Deep Learning}

Recent advances in graph deep learning provide promising opportunities to unify simplification and aggregation within a common computational framework. Graph-based representations are well suited for map generalization, as geographic entities such as buildings, roads, and rivers can be modeled as nodes and edges in a spatial graph. In this context, node regression can be applied to predict attributes of geographic objects (e.g., building size, density, or significance), which directly supports decisions about whether features should be simplified, retained, or eliminated. Similarly, link prediction allows the inference of missing or potential relationships between spatial entities, enabling more informed aggregation of adjacent polygons or functionally related areas.

Graph Neural Networks (GNNs) have proven effective in such tasks, as they capture both local and global patterns through message passing and hierarchical representation learning \citep{wu2021gnnsurvey, zhang2018linkgnn}. For example, subgraph-based GNN approaches can learn heuristics for link prediction by encoding context-specific structural patterns \citep{Zhang2018}, while line graph transformations provide more refined analysis of relationships among features \citep{Cai2021}. These advances suggest that GNNs can move beyond handcrafted rules by automatically learning how spatial entities should be simplified or aggregated based on both geometry and semantics.

Nevertheless, most existing applications of node regression and link prediction have been developed for generic social or information networks rather than spatial networks. Their potential for map generalization remains underexplored, especially in integrating simplification and aggregation tasks into a joint framework. Developing models that leverage graph-based deep learning to capture spatial dependencies, semantic relationships, and cartographic rules represents a key research frontier, which this study aims to advance.

\section{Methodology}
\subsection{Data Preparation}
The dataset used in this study consists of building footprint maps of Stuttgart at two scale pairs: 1:10,000--1:25,000 and 1:10,000--1:15,000. For the 1:10,000--1:25,000 scale pair, 91,162 one-to-one correspondences and 7,875 one-to-many relationships are identified. The 1:10,000--1:15,000 scale pair contains 110,084 one-to-one correspondences and 2,934 one-to-many relationships.

As simplification and aggregation operations are predominantly associated with one-to-many relationships, subsequent analyses focus on these cases. To facilitate the learning of simplification and aggregation operators, the selected one-to-many relationships are further transformed into graph-based representations, enabling the application of graph neural network models.

\begin{figure}[!htbp]
	\centering
		\includegraphics[scale=.35]{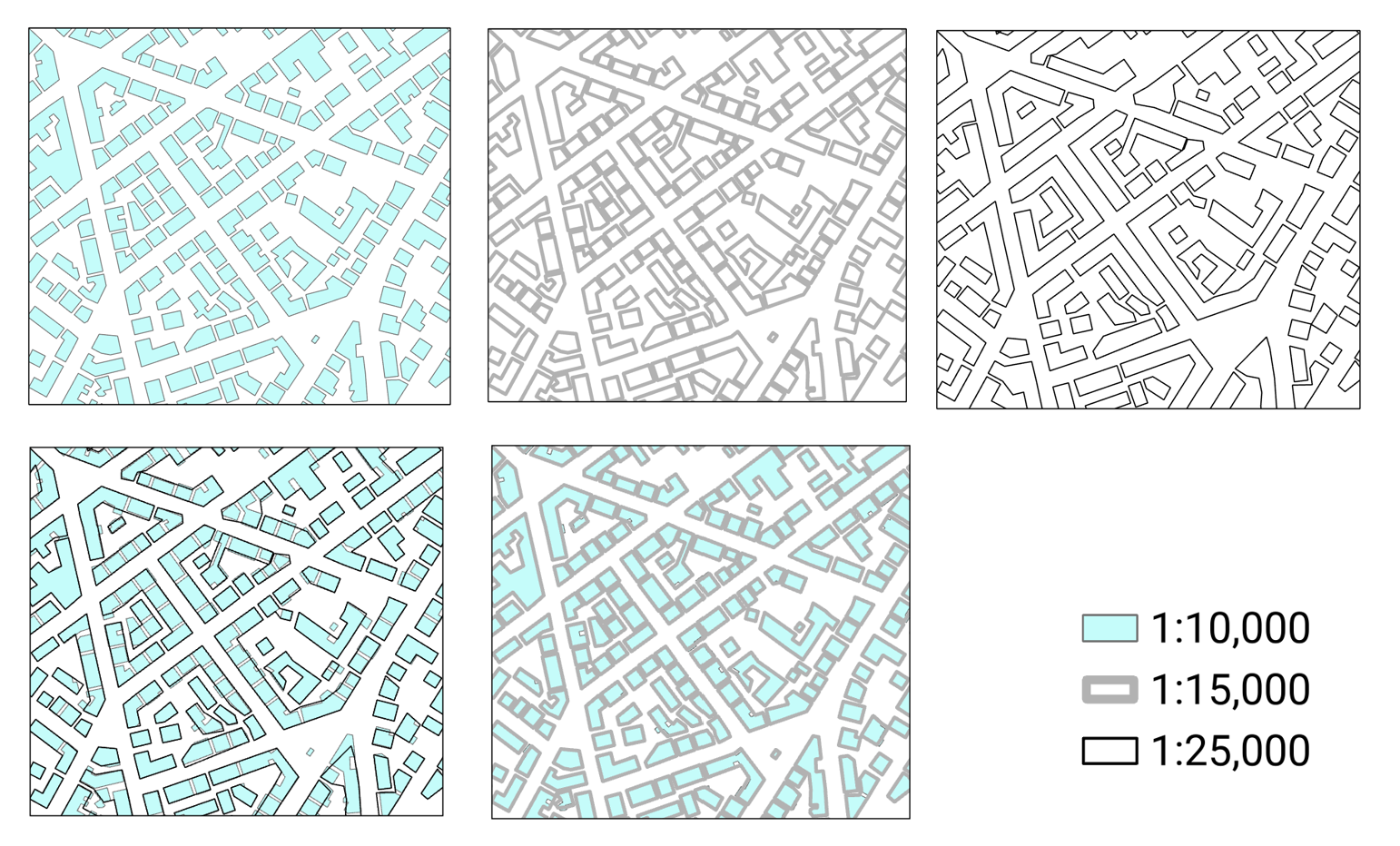}
	\caption{ Dataset used for learning the simplification and aggregation operators: blue for buildings at a map scale of 1:10,000 and polygons with black borders indicating the buildings at a map scale of 1:25,000 and grey borders indicating scale of 1:15,000 after map generalization (WGS 1984 UTM Zone 32N)}
\label{FIG:1}
\end{figure}

To improve computational efficiency in graph construction, feature extraction, and model training, the original coordinate system is transformed into a local coordinate system. This transformation preserves the relative spatial relationships among vertices within each building while providing a more compact and numerically stable representation for learning.

In the resulting graph representation, polygon vertices are mapped to graph nodes, and edges are established between consecutive vertices to encode polygonal boundaries. Node connectivity is subsequently represented using an adjacency matrix, which serves as the structural input for graph-based learning.

\begin{figure}[!htbp]
	\centering
		\includegraphics[scale=.4]{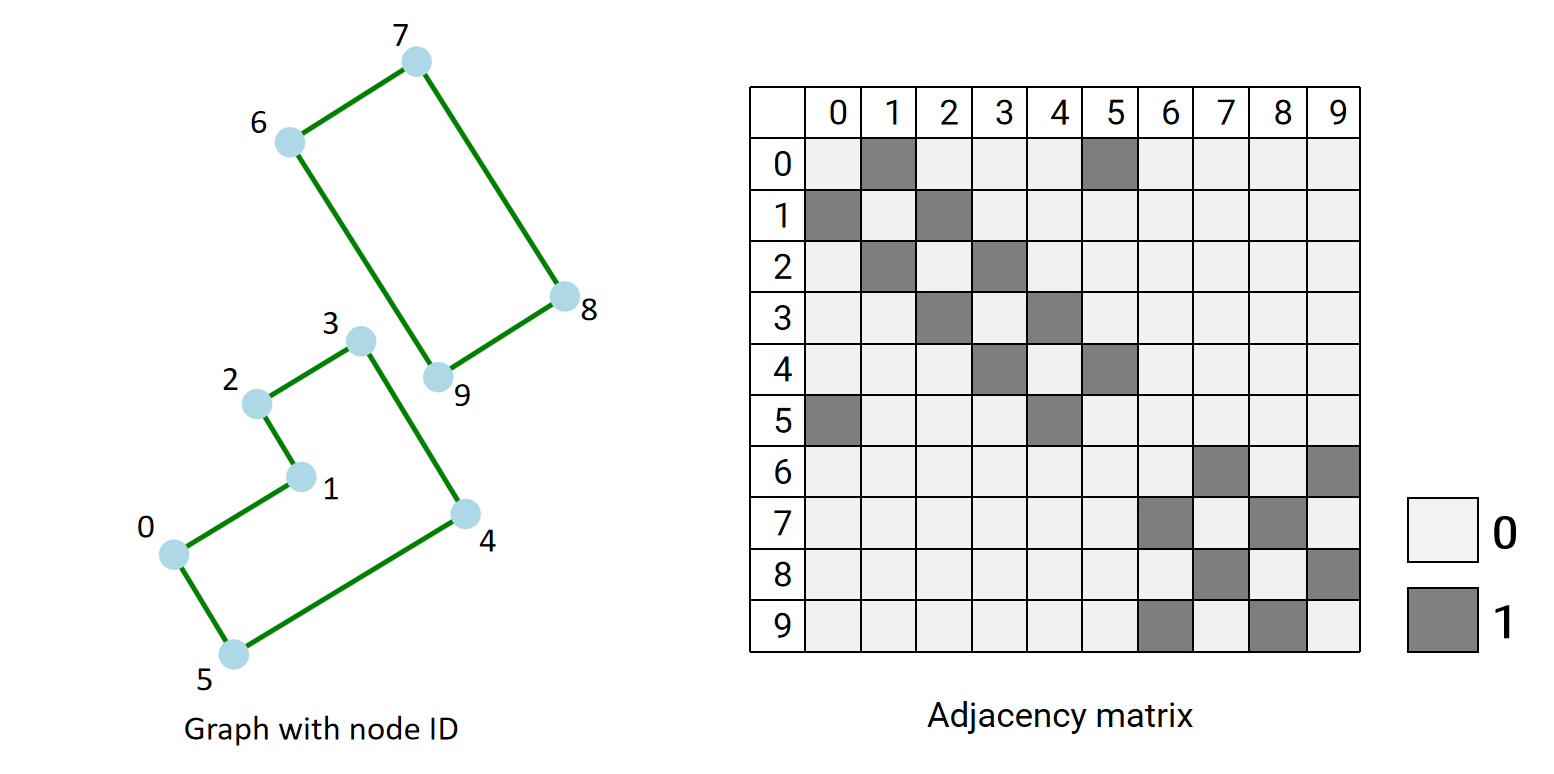}
	\caption{Graph Representation with Node IDs and Corresponding Adjacency Matrix}
	\label{FIG:2}
\end{figure}

Ground truth data are generated by transforming the graph constructed from building polygons at the 1:10,000 scale (denoted as G-geb10) to approximate the corresponding graphs at coarser or finer scales, specifically 1:25,000 (G-geb25) and 1:15,000 (G-geb15). For clarity, the transformation procedure is described using the 1:10,000--1:25,000 scale pair, while the same principles are applied to other scale combinations.

This transformation is achieved through a combination of node position adjustment and edge modification applied to G-geb10. Accordingly, the ground truth used for supervised learning consists of two components: (1) node displacement vectors derived from positional changes in G-geb10, and (2) an updated adjacency matrix reflecting the target topological structure. These components jointly define the expected outputs of the predictive model for node movement and link prediction.

\begin{figure}[h!]
	\centering
		\includegraphics[scale=.6]{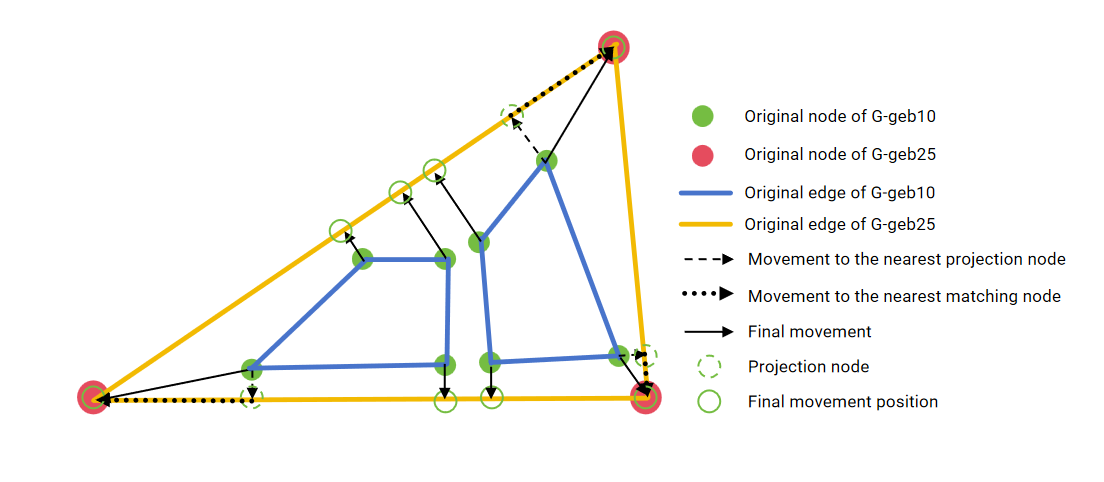}
	\caption{Node Movement and Projection Process for Ground Truth Calculation }
	\label{FIG:5}
\end{figure}

For node displacement generation, the objective is to minimize geometric movement while preserving spatial topology. Each node in G-geb10 is projected onto the edges of G-geb25 to determine its target position. To simplify correspondence establishment, every node in G-geb10 is first associated with its nearest edge in G-geb25. If the projected location coincides with an existing node in G-geb25, the G-geb10 node is relocated to that position; otherwise, it is moved to the nearest projection point along the edge. This procedure ensures that the resulting node positions closely follow the geometry of G-geb25 while introducing minimal distortion to the original configuration of G-geb10, as illustrated in Figure~\ref{FIG:5}.

After node positions are updated, the connectivity of G-geb10 is modified to ensure consistency with the topology of G-geb25. This step produces the target adjacency structure required for supervising the link prediction task.

In the second step, the edge structure of G-geb10 is updated to conform to the topology of G-geb25. All original edges in G-geb10 are first removed. The repositioned nodes obtained in the previous step are then connected sequentially to form a single closed cycle, ensuring that each node has a degree of two. 

\begin{figure}[h!]
	\centering
		\includegraphics[scale=.6]{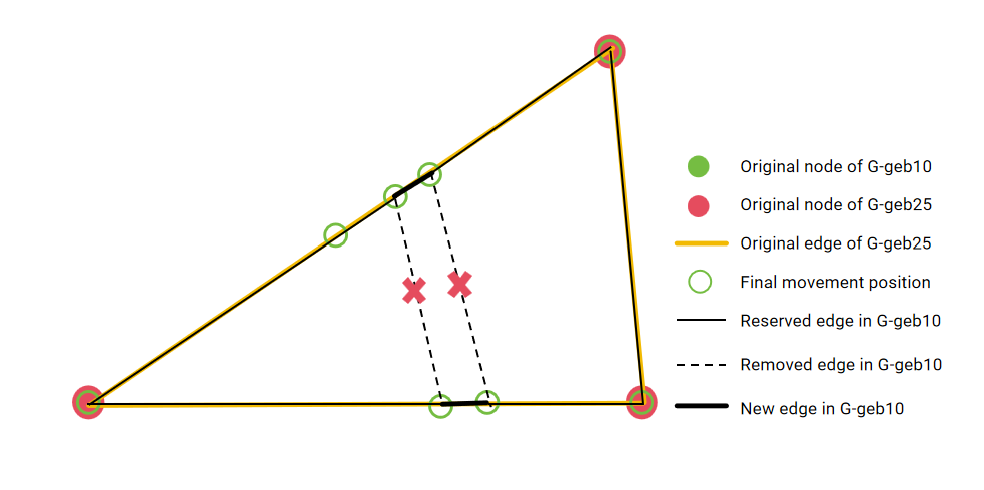}
	\caption{Edge Update Process for Ground Truth Calculation}
	\label{FIG:6}
\end{figure}

Any remaining nodes in G-geb10 that were not included in the initial cycle are subsequently projected onto the newly formed edges and inserted accordingly, preserving their spatial order along the boundary. This procedure enforces topological consistency by preventing the creation of internal connections or isolated nodes, as illustrated in Figure~\ref{FIG:6}.

The preceding figures illustrate the principles of node displacement and edge restructuring used in ground truth generation. Based on these principles, the following examples apply the proposed procedure to representative datasets to derive the corresponding ground truth. For each example, the resulting node displacement vectors and the updated adjacency matrix of G-geb10 are reported, providing explicit supervision targets for node movement and link prediction tasks.

\begin{figure}[h!]
	\centering
		\includegraphics[scale=.5]{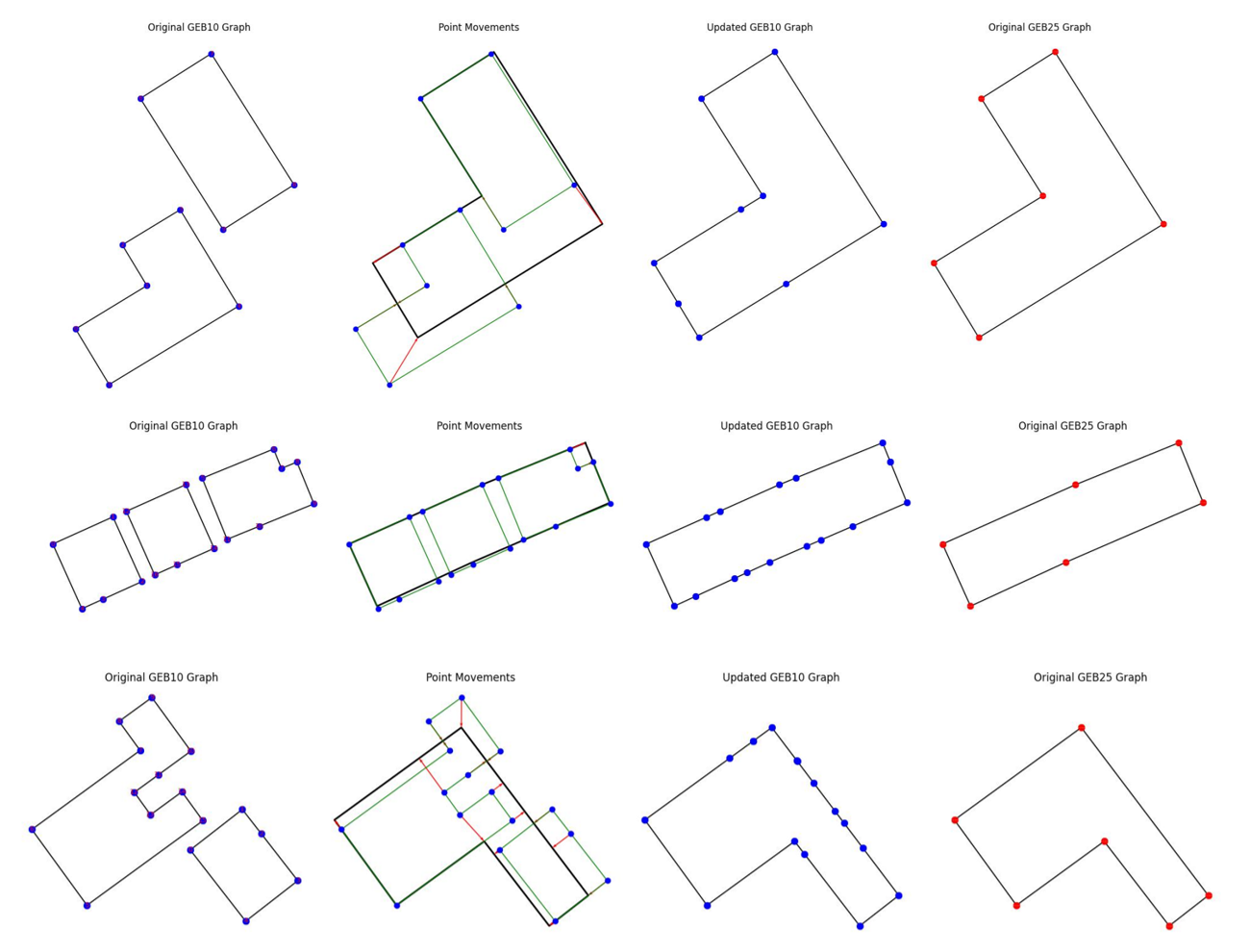}
	\caption{The process of training data preparation: (1) original buildings at map scale 1:10,000; (2) vertex movement
illustrated in red with adjacency updates; (3) aggregated building at target map scale; (4) buildings after removing redundant vertices}
	\label{FIG:1}
\end{figure}

\begin{table}[h!]  
\caption{Node Movement Data}
\label{table:nodemovement}
\begin{tabularx}{.5\textwidth}{@{} l S[table-format=3.3] S[table-format=3.3] S[table-format=1.4] S[table-format=1.4] @{}}
\toprule
\textbf{Node ID} & \textbf{Original X} & \textbf{Original Y} & \textbf{Delta X} & \textbf{Delta Y} \\
\midrule
0 & 193.419 & 193.863 & 3.6284 & 2.1949 \\
1 & 199.619 & 197.633 & -2.5801 & -1.5608 \\
2 & 197.479 & 201.173 & 0 & 0 \\
3 & 202.519 & 204.233 & -0.0170 & 0.0276 \\
4 & 207.599 & 195.853 & -1.1831 & 1.9244 \\
5 & 196.339 & 189.033 & 2.49 & 4.08 \\
6 & 199.049 & 213.903 & 0 & 0 \\
7 & 205.219 & 217.803 & -0.0020 & 0.0031 \\
8 & 212.419 & 206.403 & 0.2329 & 0.1471 \\
9 & 206.259 & 202.503 & -1.85 & 2.93 \\
\bottomrule
\end{tabularx}
\end{table}

\begin{figure}[h!]
	\centering
		\includegraphics[scale=.4]{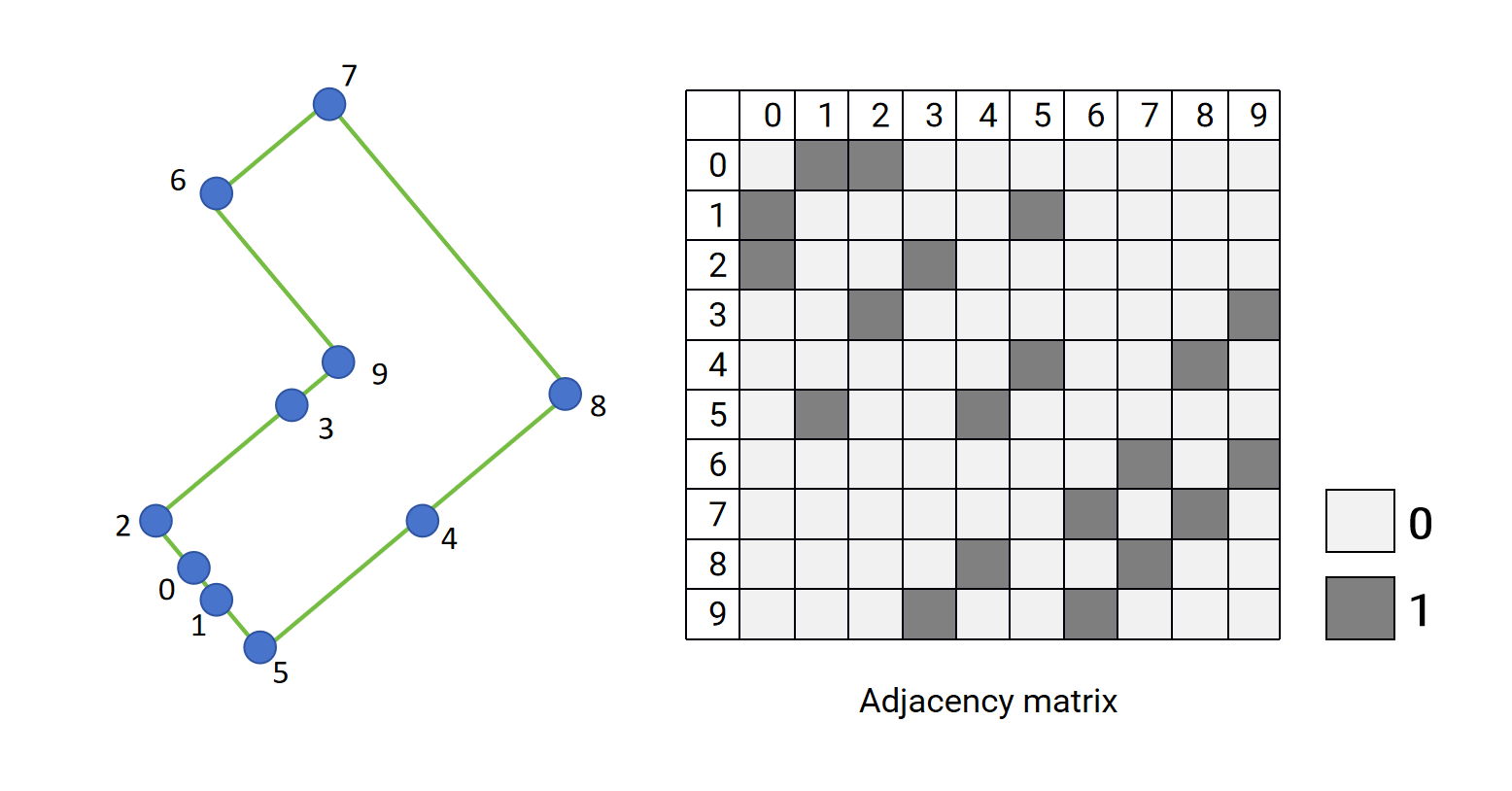}
	\caption{ Updated G-geb10 and Adjacency Matrix (Corresponding to the Original State in Figure 4)}
	\label{FIG:1}
\end{figure}

Following pre-processing, the dataset is divided into training, validation, and test subsets 
to support model development and evaluation. The allocation ratios are summarized 
in Table~\ref{tab:dataset_distribution}.

\begin{table}[htbp]
\centering
\caption{Dataset Split at Different Scales Group}
\label{tab:dataset_distribution}
\begin{tabular}{@{}lcc@{}}
\toprule
 & \textbf{1:10,000 and 1:15,000} & \textbf{1:10,000 and 1:25,000} \\ \midrule
Training set     & 74\% - 5500  & 66\% - 1900 \\
Validation set   & 13\% - 1000  & 17\% - 500  \\
Test set         & 13\% - 1000  & 17\% - 500  \\
\bottomrule
\end{tabular}
\end{table}

\subsection{Feature Engineering}

In graph-based learning, node features provide the numerical representation required to encode geometric and structural information associated with each graph element. These features are organized into a feature matrix, which, together with the adjacency matrix, forms the primary input to graph neural network models. While the adjacency matrix encodes topological connectivity, the feature matrix describes local geometric properties, enabling the model to jointly exploit spatial attributes and relational structure.

\begin{figure}[h!]
	\centering
		\includegraphics[scale=.55]{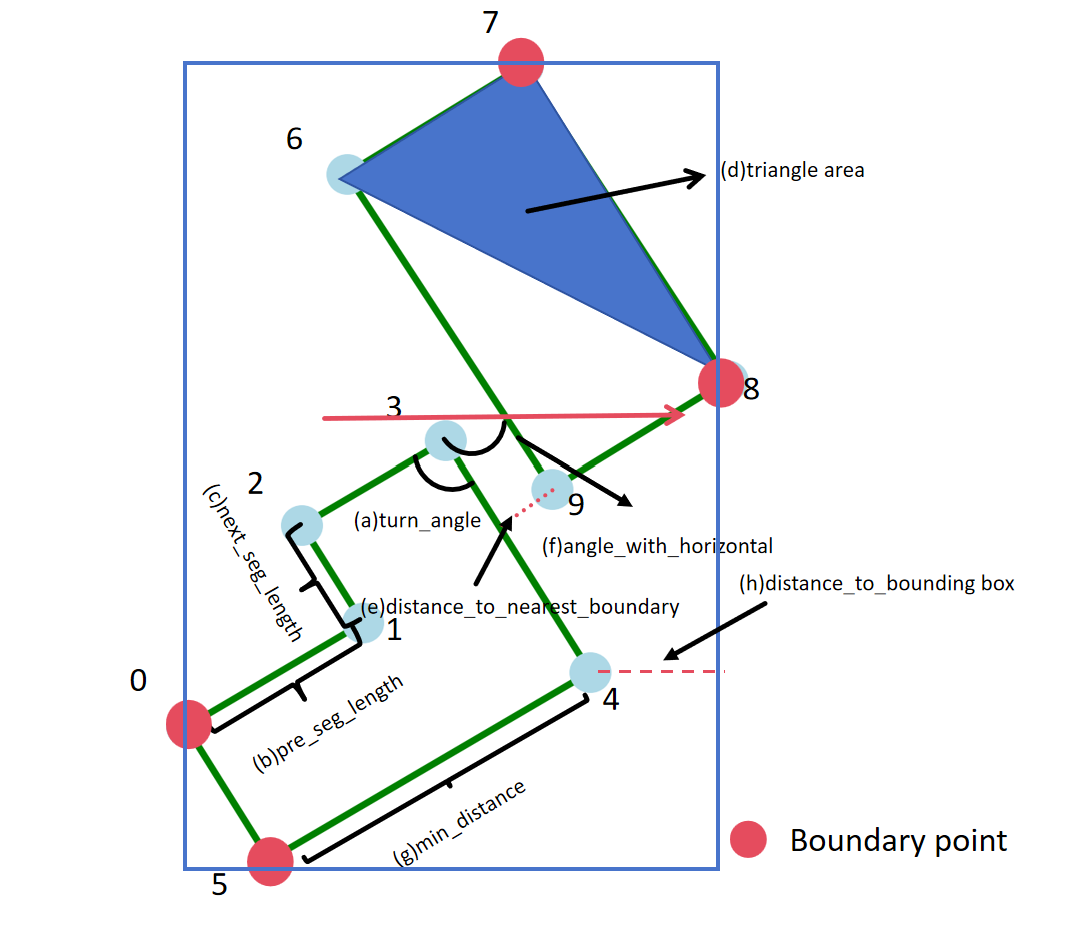}
	\caption{Geometric Features of G-geb10}
	\label{FIG:9}
\end{figure}

In this study, graphs are constructed from building polygons, where geometric relationships among vertices and edges play a central role in map generalization. Accordingly, the feature matrix is designed to capture both intrinsic and relative geometric characteristics. Intrinsic features describe properties of individual nodes and edges, such as edge length, interior angle, and local area-related attributes. Relative features encode spatial relationships between neighboring nodes and edges, including angular differences and distance-based measures. Figure~\ref{FIG:9} illustrates the computation and composition of the adopted feature sets.

For intrinsic features:

1. Turn angle($\text{loc\_turn\_angle}$): 
\[
u = \text{pos\_prev} - \text{pos\_node}, \quad v = \text{pos\_next} - \text{pos\_node}.
\]

where u and v are vectors pointing from the current node to the previous and next nodes respectively.The angle between vectors \(\mathbf{u}\) and \(\mathbf{v}\) is given by:
\[
\text{angle} = \arccos \left( \frac{\mathbf{u} \cdot \mathbf{v}}{\|\mathbf{u}\| \|\mathbf{v}\|} \right)
\]

2. Segment Lengths (\(\text{pre\_seg\_length}\) and \(\text{next\_seg\_length}\)): 
\begin{align*}
\text{pre\_length} &= \| \mathbf{u} \| = \sqrt{u_x^2 + u_y^2} \\
\text{next\_length} &= \| \mathbf{v} \| = \sqrt{v_x^2 + v_y^2}
\end{align*}
Here, \( u_x \) and \( u_y \), as well as \( v_x \) and \( v_y \) are the \( x \) and \( y \) components of vectors \( \mathbf{u} \) and \( \mathbf{v} \) respectively.The average of these two segment lengths is defined as the Local Segment Length:
\[
\text{Local Segment Length} = \frac{\text{pre\_length} + \text{next\_length}}{2}
\]

3. Triangle Area \(\text{loc\_tri\_area}\), is calculated using half the modulus of the cross product of vectors \( \mathbf{u} \) and \( \mathbf{v} \):
\[
\text{area} = 0.5 \cdot |\mathbf{u} \times \mathbf{v}|
\]

For relative features, this study has Distance to Nearest Boundary Point, Angle to Horizon, and Distance to Boundary. 

1. The minimum distance from the position node to the nearest boundary point is given by:
\begin{multline}
\text{min\_distance} = \min_{\text{boundary\_point} \in \text{boundary\_points}} \\
|\text{pos\_node} - \text{boundary\_point}|
\end{multline}

2. The angle to the horizon, based on the differences in x and y coordinates, is given by:
\[
\text{angle\_to\_horizon} = \arctan2(\text{dy}, \text{dx})
\]

After feature computation, feature standardization is applied to ensure numerical stability and efficient model training. Standardization is performed using a z-score transformation, which preserves the relative information within each feature while bringing different feature dimensions to a comparable scale. This process facilitates stable optimization and accelerates convergence during training.

To prevent data leakage, the mean and standard deviation required for standardization are computed exclusively from the training dataset. These statistics are then applied consistently to the validation and test datasets. Using information from validation or test data to estimate normalization parameters would introduce leakage and potentially compromise the generalization performance of the trained models.

In addition to feature normalization, a constrained Delaunay triangulation is constructed to support spatial modeling and post-processing. Delaunay triangulation provides an efficient means of organizing vertices into a triangular mesh while preserving local neighborhood relationships. In this study, building edges are treated as constraints to ensure that all original polygon edges are retained within the triangulation.

To simplify boundary handling and ensure complete triangulation, an enclosing bounding box defined by four auxiliary vertices is introduced to encompass all points. This approach mitigates ambiguities associated with boundary vertices, which may otherwise lack sufficient neighboring points to form valid triangles. By allowing boundary points to connect with the bounding box vertices, a closed and complete triangulation is obtained.

\begin{figure}[h!]
	\centering
		\includegraphics[scale=.55]{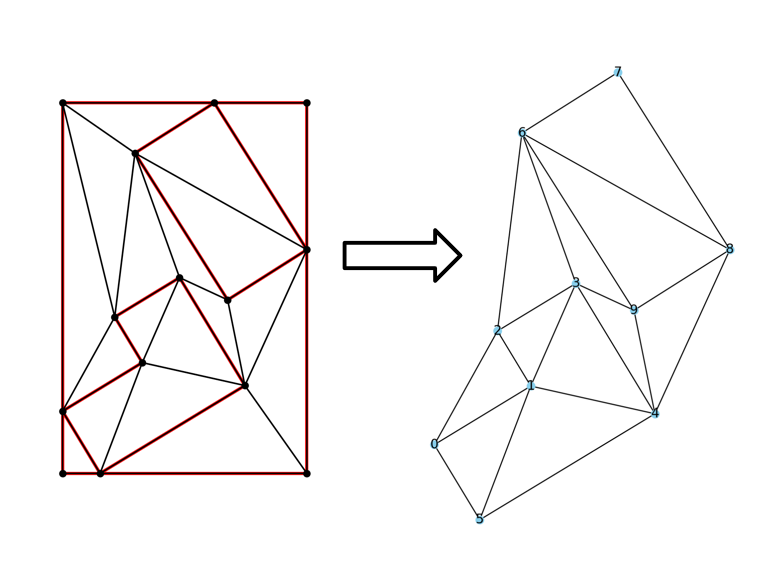}
	\caption{Building Edge Constraint Delaunay Triangulation}
	\label{FIG:10}
\end{figure}

The use of a bounding box also improves triangle quality by reducing the occurrence of elongated or sharply angled triangles, which can negatively affect numerical stability. The resulting triangulation provides a regular and well-conditioned spatial structure for subsequent analysis, as illustrated in Figure~\ref{FIG:10}.

\subsection{Model Architecture and Training Strategy}

All models were implemented using PyTorch. A modular graph neural network (GNN) architecture was adopted to facilitate flexible experimentation with different convolution operators and training strategies. Residual connections and batch normalization layers were incorporated to improve gradient propagation and stabilize training, particularly for deeper graph convolutional stacks.

To learn node-level representations, three representative GNN architectures were employed as backbone encoders: Graph Convolutional Networks (GCN), Graph Attention Networks (GAT), and GraphSAGE. Each graph convolution layer aggregates information from neighboring nodes to update node embeddings, followed by batch normalization to improve convergence stability. The final node embeddings are subsequently passed to task-specific output heads for link prediction and node movement estimation.

\paragraph{Link Prediction for Aggregation.}
Building aggregation is formulated as a link prediction problem on the graph, where each entry in the adjacency matrix can assume one of three discrete states: deletion ($-1$), retention ($0$), or addition ($+1$) of an edge. These states are encoded using one-hot vectors and optimized using a cross-entropy loss.

To reflect the asymmetric and irreversible nature of aggregation operations, structural constraints are imposed on the prediction space. Specifically, edges that do not exist in the original graph are restricted to either be added or ignored, while existing edges are restricted to either be retained or deleted. By eliminating infeasible predictions, this constraint reduces ambiguity in the learning process and improves training efficiency.

Predicted edge states are reshaped to the appropriate dimensions and transformed into class probabilities using the Softmax function. To further restrict predictions to geometrically plausible regions, a triangular mesh adjacency matrix derived from Delaunay triangulation is used as a spatial mask. This mask confines edge modifications to local neighborhoods and reduces the occurrence of spurious internal connections.

\paragraph{Node Movement for Simplification.}
Node movement is modeled as a regression task to support building simplification. In addition to the original geometric features, positional features such as inter-node distances and angular relationships are dynamically computed during forward propagation and concatenated with node embeddings. This design enables the model to better capture spatial context and local geometric variation.

Node displacement is optimized using a mean squared error (MSE) loss between predicted and target node positions. To preserve the geometric and topological integrity of building outlines after node movement, auxiliary losses based on edge length consistency and angular deviation within the triangular mesh are incorporated. These regularization terms encourage shape preservation and reduce topological artifacts introduced during prediction.

\paragraph{Multi-task Loss Balancing.}
During training, the magnitude of the node movement loss was observed to be approximately two orders of magnitude larger than that of the link prediction loss. To balance their contributions during joint optimization and prevent either task from dominating the learning process, the total loss is defined as:
\begin{equation}
\mathcal{L}_{\text{total}} = \mathcal{L}_{\text{node}} + 100 \times \mathcal{L}_{\text{edge}}.
\end{equation}
This weighting factor was empirically determined to ensure stable convergence across both tasks.

\begin{figure}[h!]
	\centering
		\includegraphics[scale=.4]{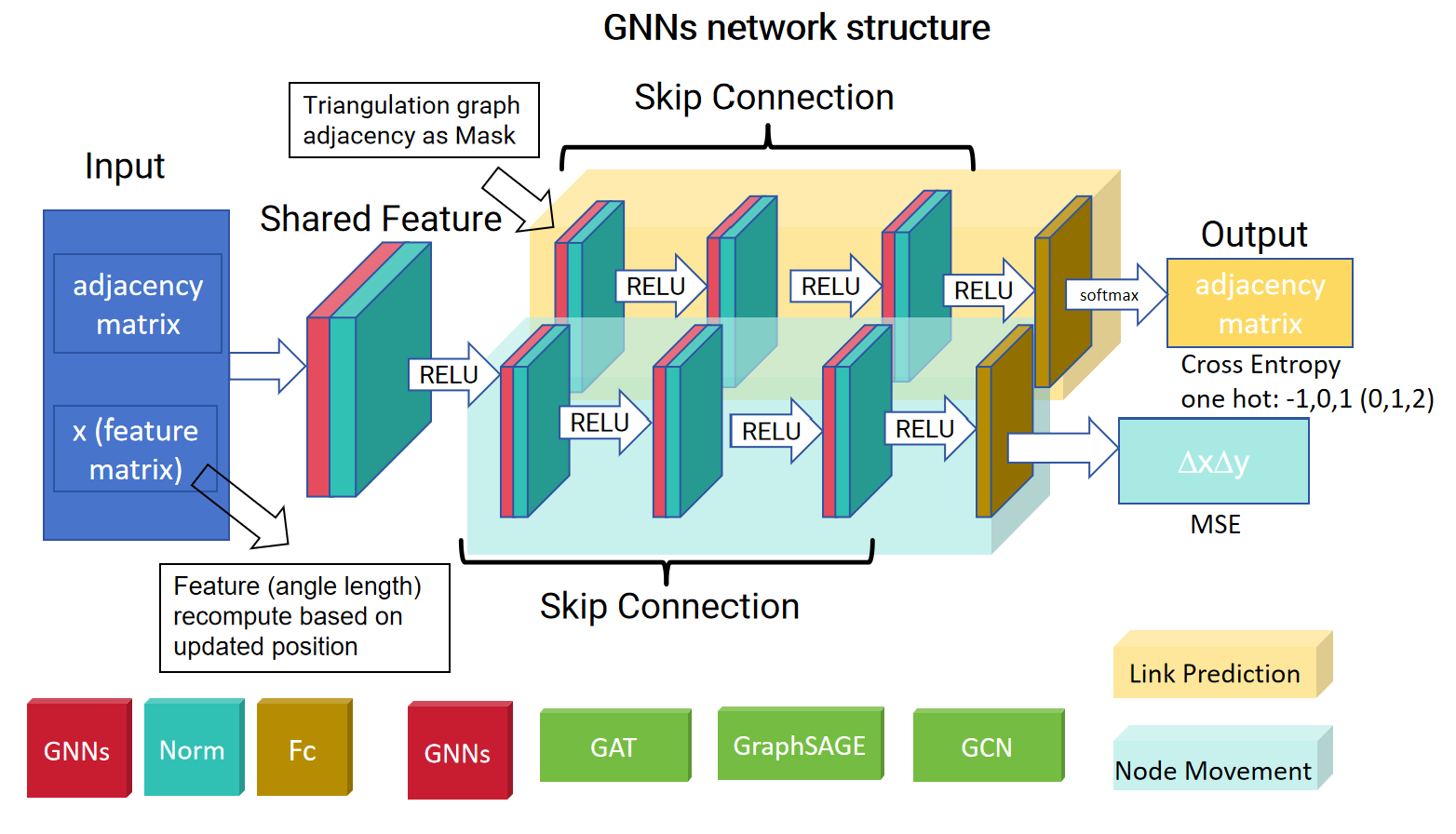}
	\caption{Graph Neural Network (GNN) Model Architecture for Link Prediction and Node Movement}
	\label{FIG:1}
\end{figure}

\begin{table}[h!]
\centering
\caption{Hyperparameters for Node Movement and Link Prediction Tasks}
\label{tab:hyperparameters}
{\footnotesize 
\begin{tabularx}{\columnwidth}{@{} >{\hsize=.3\hsize}X >{\hsize=.35\hsize}X >{\hsize=.35\hsize}X @{}}  
\toprule
\textbf{Hyperparameter} & \textbf{Node Movement} & \textbf{Link Prediction} \\
\midrule
Input Channels & Original features + 2 & Original features \\
Output Channels & 2 & 40 \\
Hidden Layers& 64 & 64 \\
Batch Size & 16 & 16 \\
Learning Rate & 5e-4 & 5e-4 \\
Optimizer & Adam & Adam \\
\bottomrule
\end{tabularx}
}
\end{table}

\subsection{Hamiltonian Cycle-based Post-processing for Aggregation}

Predicted adjacency structures obtained from graph neural network models do not always satisfy the topological requirements of valid building outlines. In particular, predicted graphs may contain superfluous internal connections or missing edges, resulting in node degrees deviating from two and preventing the formation of closed polygonal cycles. Such structural inconsistencies limit the direct usability of link prediction outputs for map generalization.

To address this issue, a post-processing step based on Hamiltonian cycle construction is introduced to enforce structural validity. From a graph-theoretic perspective, a valid aggregated building outline can be represented as a simple cycle in which each node has degree two and the path forms a closed loop. The Hamiltonian cycle formulation seeks to identify a cycle that visits each node exactly once and returns to the starting node, thereby eliminating redundant internal edges while preserving global connectivity.

In the proposed framework, Hamiltonian cycle construction is applied to predicted graphs that contain sufficient connectivity but exhibit topological redundancy. Edge weights are defined based on geometric criteria, such as edge length and angular deviation, enabling the selection of a geometrically plausible cycle among multiple candidates. This process removes spurious internal connections and restores a consistent polygonal structure suitable for aggregation results.

It is important to note that Hamiltonian cycle-based post-processing is not universally applicable. If the predicted graph lacks essential connections or contains isolated nodes, a valid Hamiltonian cycle cannot be constructed. Consequently, this post-processing strategy primarily serves as a corrective mechanism for structurally recoverable predictions rather than a complete solution for all aggregation errors. In addition to correction, the success or failure of Hamiltonian cycle construction also provides an implicit measure of the structural soundness of the original link prediction outputs.

\section{Experiments}
\subsection{Comparative Analysis across Models and Datasets}

This study conducted comparative experiments to evaluate the performance of three graph neural network architectures—GCN, GAT, and GraphSAGE—on both link prediction and node movement tasks. Experiments were performed on multiple dataset groups, with the 1:10,000--1:25,000 scale pair presented here as a representative example. Model performance was assessed using accuracy for the link prediction task and mean squared error (MSE) for the node movement task, together with training and validation loss curves to analyze convergence behavior and generalization.

\begin{figure}[h!]
	\centering
		\includegraphics[scale=.35]{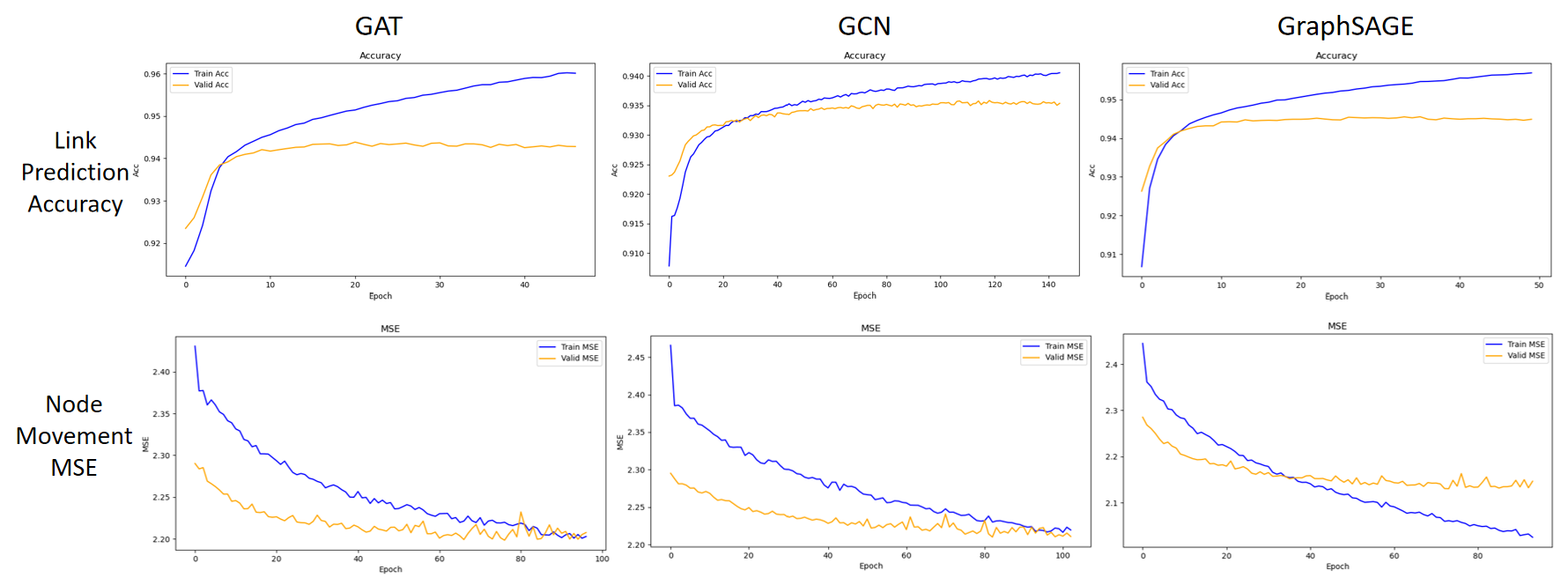}
	\caption{Training and Validation Performance of GNN Models for Link Prediction Accuracy, and Node Movement Loss in 1:10,000 and 1:25,000 Datasets}
	\label{FIG:12}
\end{figure}

Figure~\ref{FIG:12} illustrates the training and validation performance of the three models. For the link prediction task, GraphSAGE consistently achieves higher accuracy on both training and validation sets, particularly during the early stages of training. This suggests that GraphSAGE is more effective in capturing structural relationships relevant to aggregation. In contrast, GAT and GCN exhibit more gradual improvements in accuracy, with relatively similar convergence behavior.

From the loss curves, GCN shows a rapid decrease in training loss at the beginning of optimization, indicating faster initial adaptation. However, its validation loss exhibits larger fluctuations compared to the other models. In comparison, GraphSAGE and GAT demonstrate smoother loss trajectories, with GraphSAGE achieving the most stable validation performance, which may indicate improved robustness and generalization to unseen data.

For the node movement task, GCN achieves a sharper reduction in training error, while GAT and GraphSAGE show more moderate but stable decreases in both training and validation losses. This observation suggests that GCN may adapt more aggressively to node displacement patterns during early training, whereas GAT and GraphSAGE emphasize more stable learning dynamics over longer training horizons. Despite these differences, all three models exhibit notable challenges in accurately predicting large node displacements, highlighting the inherent difficulty of the node movement task.

Overall, the comparative results indicate that GraphSAGE provides the most balanced performance across both tasks, combining relatively high link prediction accuracy with stable convergence behavior. GAT follows closely, particularly in terms of validation stability, while GCN demonstrates strong initial learning capability but comparatively weaker generalization.

\begin{table}[htbp]
\centering
\caption{Performance Metrics for Different Graph Models}
\label{tab:performance_metrics}
{\footnotesize
\begin{tabular}{@{}lcccccc@{}}
\toprule
\textbf{Model} & \multicolumn{2}{c}{\textbf{GAT}} & \multicolumn{2}{c}{\textbf{GCN}} & \multicolumn{2}{c}{\textbf{GraphSAGE}} \\
\cmidrule(r){2-3} \cmidrule(lr){4-5} \cmidrule(l){6-7}
\textbf{Metric} & \textbf{Train} & \textbf{Valid} & \textbf{Train} & \textbf{Valid} & \textbf{Train} & \textbf{Valid} \\
\midrule
Accuracy & 0.9602 & 0.9438 & 0.9405 & 0.9358 & 0.9568 & 0.9455 \\
MSE & 2.2008 & 2.1984 & 2.2708 & 2.2579 & 2.0193 & 2.1350 \\
\bottomrule
\end{tabular}
}
\end{table}

Across both link prediction and node movement tasks, GraphSAGE demonstrates consistently strong performance, achieving higher accuracy and more stable convergence compared to GCN and GAT. In particular, GraphSAGE exhibits lower training loss and smoother validation behavior, suggesting a favorable balance between learning efficiency and generalization. GAT, while slightly behind in training performance, shows competitive validation results across both tasks, indicating robust generalization capability.

\begin{figure}[h!]
	\centering
		\includegraphics[scale=.35]{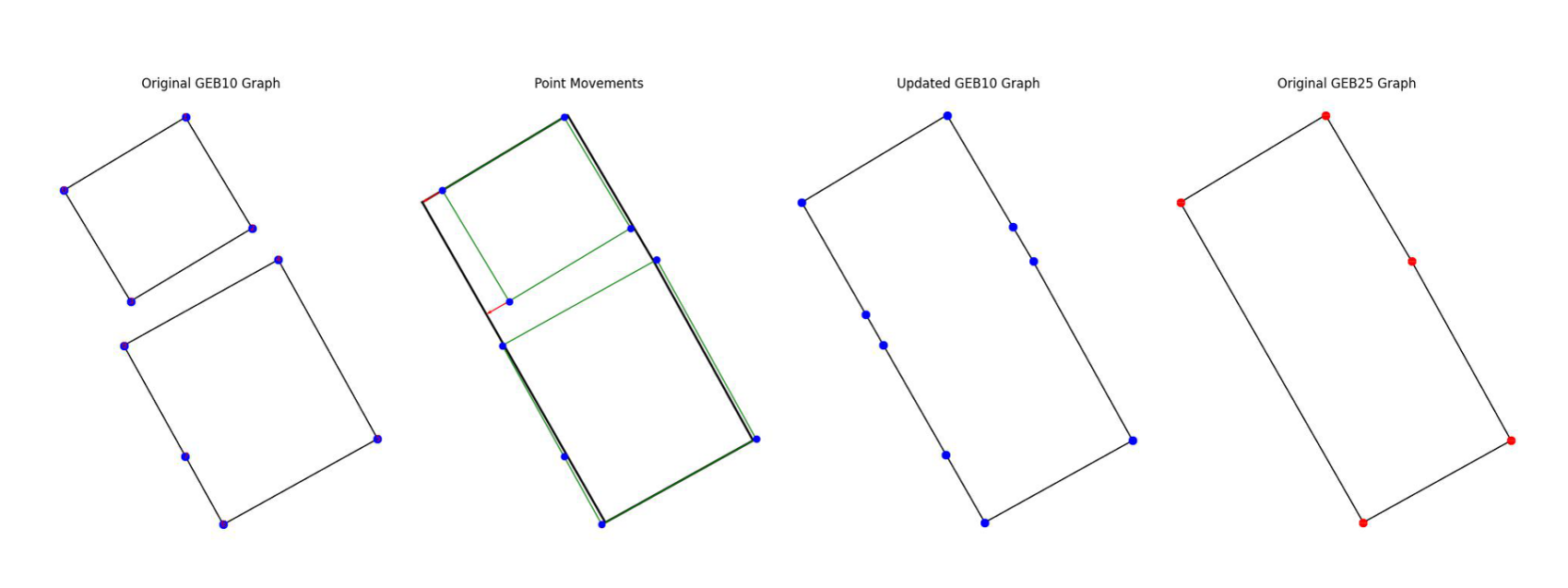}
	\caption{Ground Truth Visualization Example in Validation Datasets for Node Movement and Link Prediction Task}
	\label{FIG:13}
\end{figure}

Figure~\ref{FIG:13} presents representative results from the validation dataset, illustrating node movement predictions in terms of $\Delta x$ and $\Delta y$, as well as predicted adjacency matrices for the link prediction task. Visual comparison with the ground truth reveals that all three models are capable of capturing the overall structural patterns of building aggregation. However, noticeable discrepancies remain, particularly in predicting larger node displacements and resolving internal connections within aggregated structures. Among the three models, GraphSAGE generally produces predictions that are closer to the ground truth in terms of external outline consistency, while GAT and GCN exhibit comparable but slightly less stable results.

\begin{table}[htbp]
\centering
\caption{Comparison of Displacement Predictions}
\label{tab:displacements}
{\tiny 
\begin{tabularx}{\columnwidth}{@{}l *{8}{>{\centering\arraybackslash}X}@{}}
\toprule
& \multicolumn{2}{c}{Ground Truth} & \multicolumn{2}{c}{GAT} & \multicolumn{2}{c}{GCN} & \multicolumn{2}{c}{GraphSAGE} \\
\cmidrule(lr){2-3} \cmidrule(lr){4-5} \cmidrule(lr){6-7} \cmidrule(l){8-9}
& $\Delta x$ & $\Delta y$ & $\Delta x$ & $\Delta y$ & $\Delta x$ & $\Delta y$ & $\Delta x$ & $\Delta y$ \\
\midrule
1 & -1.48 & -0.89 & 0.1158 & -0.0525 & 0.1570 & 0.0373 & 0.4198 & 0.0092 \\
2 & 0.26 & 0.16 & 0.0720 & -0.0964 & 0.0733 & -0.3027 & 0.2587 & 0.0015 \\
3 & 0.1976 & 0.1164 & 0.4629 & 0.0077 & 0.2570 & -0.1005 & 0.2873 & 0.0127 \\
4 & -1.6832 & -0.9618 & 0.3051 & -0.0315 & 0.2659 & 0.0781 & 0.2189 & 0.0179 \\
5 & -0.22 & -0.12 & -0.4792 & -0.1218 & 0.0772 & 0.2093 & 0.0847 & 0.0284 \\
6 & -0.22 & -0.12 & -0.2283 & -0.1080 & 0.0599 & 0.1162 & -0.2795 & -0.0093 \\
7 & 0.26 & 0.15 & -0.1142 & 0.6499 & -0.0497 & -0.1539 & 0.4006 & 1.0501 \\
8 & 0.1992 & 0.1138 & 0.1376 & 0.1692 & -0.1056 & 0.3302 & -0.1095 & 0.1526 \\
9 & 0.1023 & 0.0584 & -0.2330 & -0.3537 & -0.2322 & 0.1479 & 0.1500 & -0.1889 \\
\bottomrule
\end{tabularx}
}
\end{table}

Table~\ref{tab:displacements} reports representative node movement prediction results for the three graph neural network models (GCN, GAT, and GraphSAGE) in comparison with the ground truth, illustrating discrepancies in the predicted $\Delta x$ and $\Delta y$ values. Overall, noticeable deviations between predicted and reference displacements are observed across all models, indicating that node movement prediction remains a challenging task.

Among the three models, GraphSAGE generally produces smaller errors in $\Delta x$, suggesting a relatively stronger capability in capturing horizontal node displacements. However, substantial errors are still present in the $\Delta y$ predictions, particularly for nodes exhibiting larger vertical movements. For example, in the seventh case, the predicted $\Delta y$ deviates markedly from the ground truth, highlighting the model’s difficulty in handling extreme displacement values.

These results indicate that while graph neural networks can partially learn local node movement patterns, their performance is uneven across displacement directions. In particular, all three models show limited accuracy in predicting larger or less frequent movements, underscoring the impact of data imbalance and the inherent complexity of unconstrained node displacement prediction.

\begin{figure}[h!]
	\centering
		\includegraphics[scale=.35]{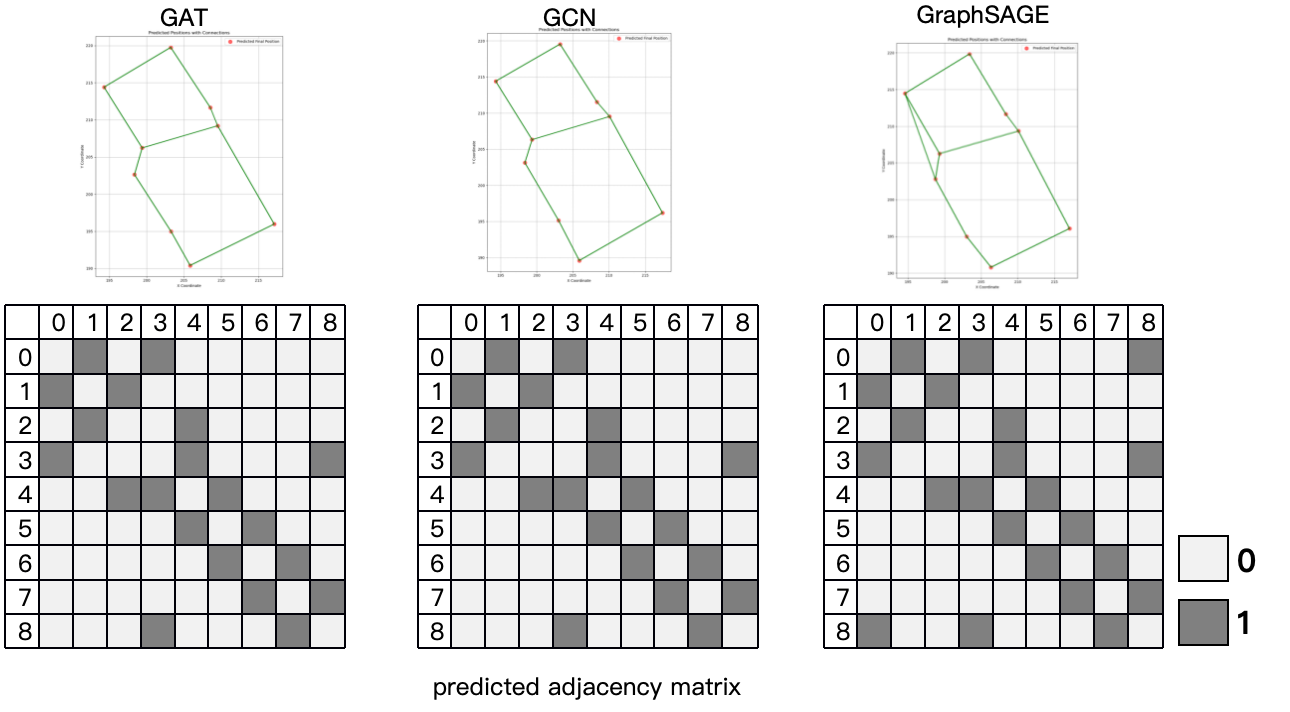}
	\caption{Predicted Adjacency Example with GAT, GCN, and GraphSAGE Models in Link Prediction Task }
	\label{FIG:14}
\end{figure}

Figure~\ref{FIG:14} illustrates the link prediction results produced by GAT, GCN, and GraphSAGE, together with the original and ground-truth adjacency matrices. In the visualization, gray cells indicate the presence of edges, while white cells represent the absence of connections. Overall, all three models are able to capture the main external structure of the aggregated buildings, indicating that graph-based link prediction can effectively support the aggregation operator.

GAT and GCN exhibit similar connection patterns in their predictions, whereas GraphSAGE produces a distinct adjacency structure. This difference suggests that GraphSAGE aggregates neighborhood information in a manner that leads to alternative structural interpretations of node relationships. While the external outlines are generally consolidated, redundant internal connections are frequently observed across all models, highlighting persistent challenges in accurately resolving internal topology during aggregation.

\begin{figure*}[h!]
    \centering
    \includegraphics[width=0.95\textwidth]{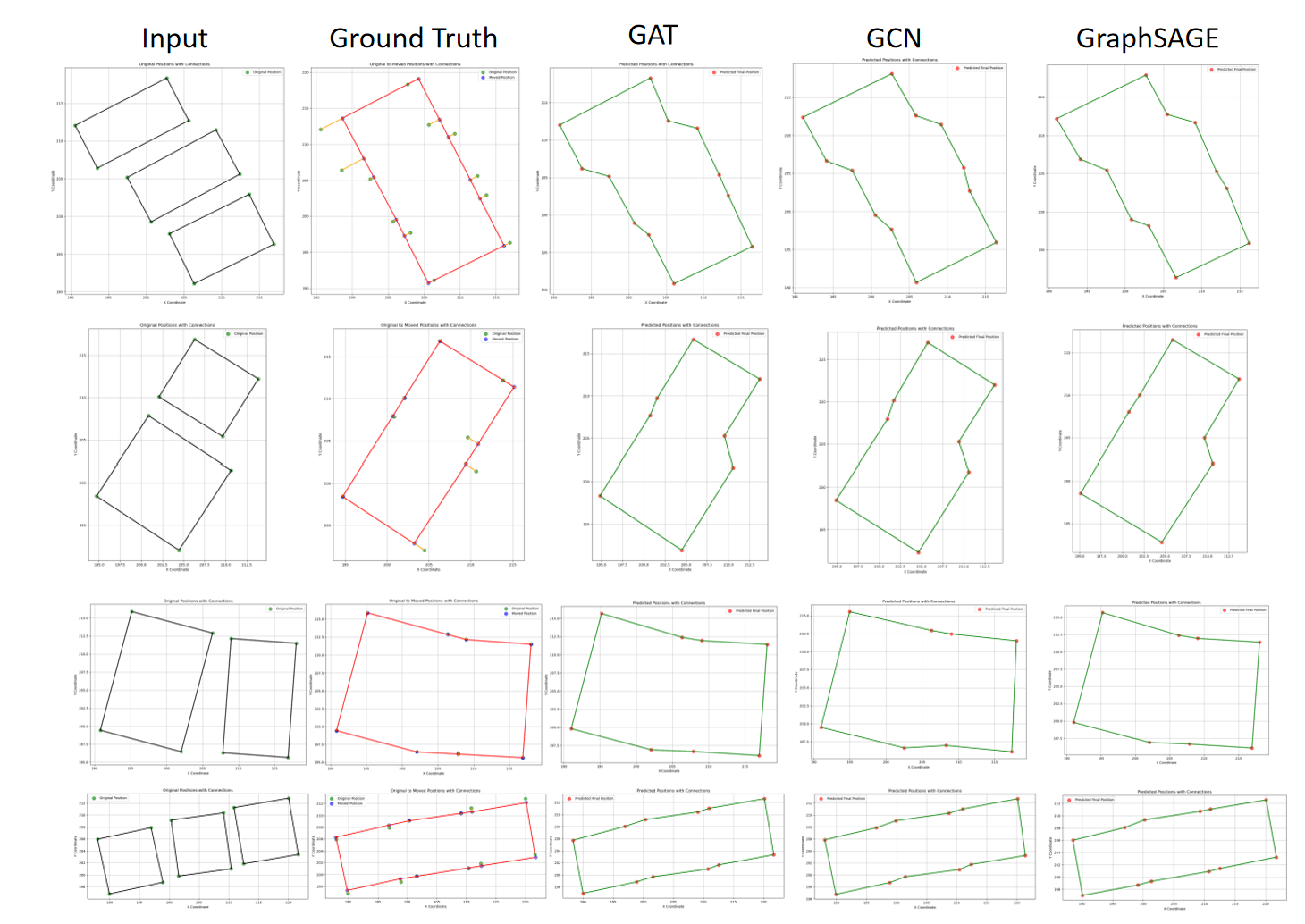}
    \caption{Prediction Comparison between different models-Part 1}
    \label{FIG:15}
\end{figure*}

Figure~\ref{FIG:15} further presents representative qualitative examples comparing the input, ground truth, and model predictions. In Part~1, predicted results closely approximate the ground truth, with only minor structural discrepancies, demonstrating that the models can successfully reconstruct aggregated building outlines in relatively simple cases. In contrast, Part~2 shows examples with substantial deviations from the reference structures. Typical errors include the prediction of spurious internal edges, missing edges required to form closed cycles, and incorrect connection patterns that disrupt topological consistency.

In addition, node movement predictions remain less reliable, particularly for nodes undergoing larger displacements. Across all models, significant errors are observed for such cases, indicating limitations in modeling large or infrequent positional changes. These observations suggest that while link prediction can capture overall aggregation patterns, accurately predicting node movement—especially under dynamic and large-scale displacements—remains a challenging problem.

\begin{figure*}[h!]
    \centering
    \includegraphics[width=0.95\textwidth]{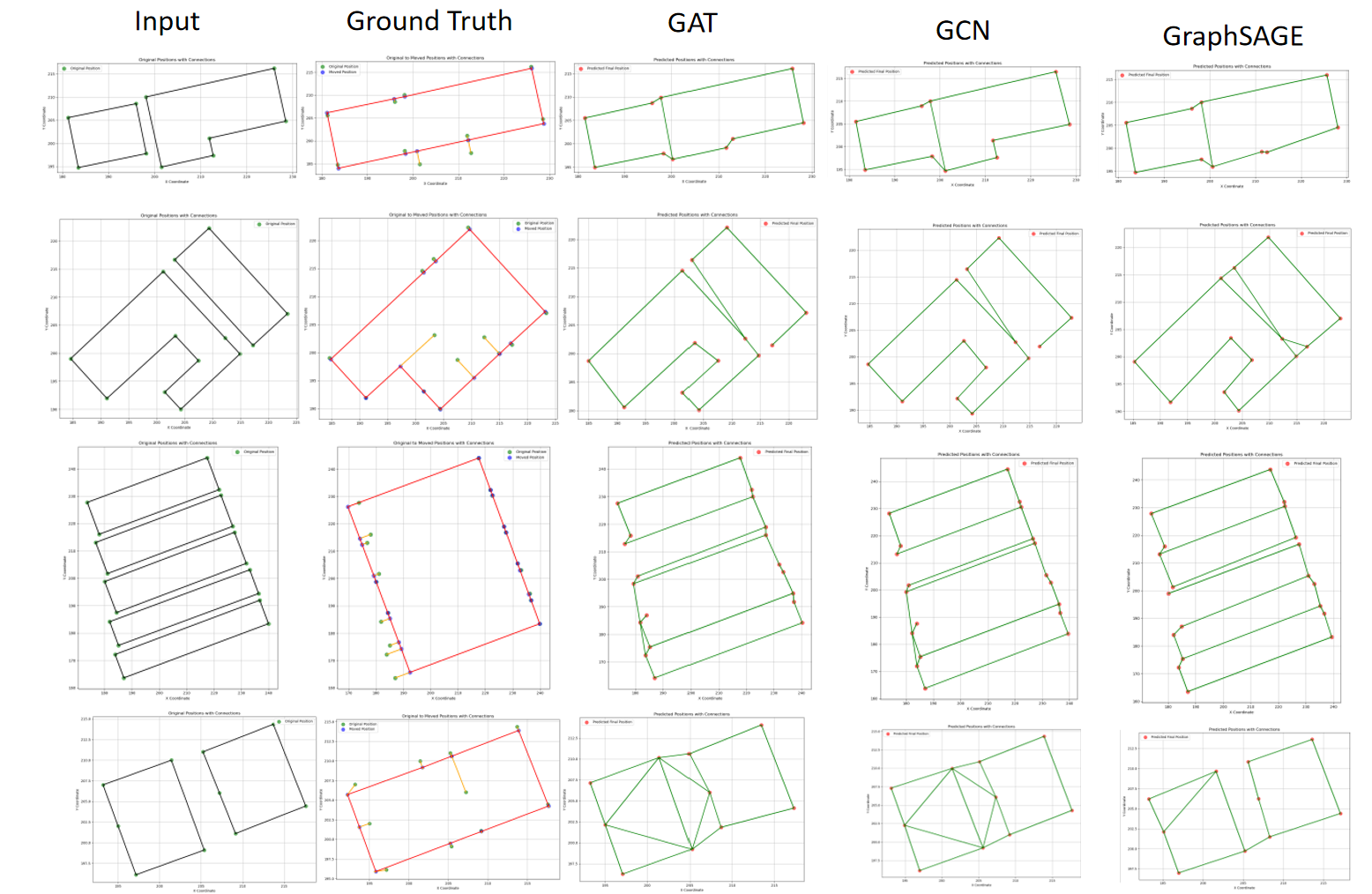}
    \caption{Prediction Comparison between different models-Part 2}
    \label{FIG:16}
\end{figure*}

In addition to cross-model comparisons, the performance of identical graph neural network architectures is further analyzed across different dataset groups to examine the influence of map scale. Specifically, experiments are conducted on two scale pairs: 1:10k--1:15k and 1:10k--1:25k.

This comparison allows an assessment of model robustness and sensitivity to varying levels of geometric detail introduced by different generalization scales. Figure~\ref{FIG:16} presents representative prediction results produced by GCN, GAT, and GraphSAGE on the two dataset groups, illustrating variations in model behavior under different scale transitions.

\begin{table}[h!]
\centering
\caption{Performance Metrics for Graph Models at Different Scales}
\label{tab:metrics}
{\tiny
\begin{tabular}{@{}lcccccc@{}}
\toprule
& \multicolumn{2}{c}{GCN} & \multicolumn{2}{c}{GAT} & \multicolumn{2}{c}{GraphSAGE} \\
\cmidrule(lr){2-3} \cmidrule(lr){4-5} \cmidrule(l){6-7}
Scale Group  & 1:15k & 1:25k  & 1:15k  & 1:25k & 1:15k & 1:25k \\
\midrule
Accuracy of Link Prediction & 0.9420 & 0.9358 & 0.9395 & 0.9438 & \textbf{0.9477} & 0.9455 \\
MSE of Node Movement & 1.0398 & 2.2579 & 1.0339 & 2.1984 & \textbf{1.0233} & 2.1350 \\
\bottomrule
\end{tabular}
}
\end{table}

Table~\ref{tab:metrics} summarizes the performance of three graph neural network models—GCN, GAT, and GraphSAGE—across two dataset groups corresponding to different scale transitions: 1:10k--1:15k and 1:10k--1:25k. The evaluated metrics include link prediction accuracy and the mean squared error (MSE) of node movement.

Across both scale groups, GraphSAGE achieves consistently high link prediction accuracy and relatively low MSE values, indicating stable performance under varying levels of generalization. GAT exhibits comparable robustness, with a slight increase in link prediction accuracy at the coarser scale, accompanied by a moderate rise in node movement error. In contrast, GCN shows a more pronounced increase in MSE at the larger scale, suggesting greater sensitivity to increased geometric simplification and larger graph transformations.

\begin{figure*}[h!]
    \centering
    \includegraphics[width=0.95\textwidth]{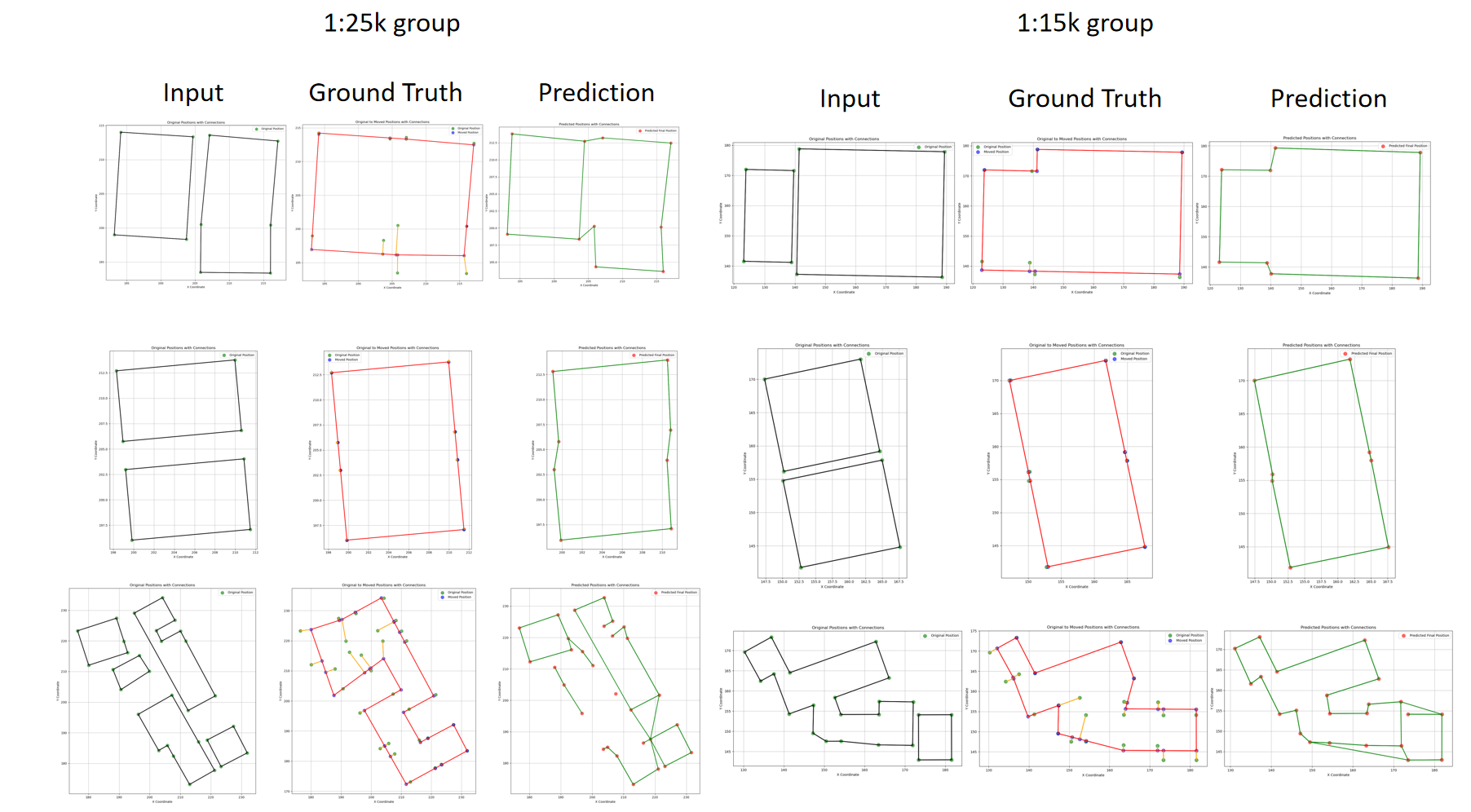}
    \caption{Prediction Comparison between different dataset groups}
    \label{FIG:17}
\end{figure*}

These results highlight differences in scalability and robustness among the evaluated models, with GraphSAGE demonstrating more consistent behavior across scale transitions. Figure~\ref{FIG:17} further illustrates qualitative prediction results using GraphSAGE as a representative example, enabling visual comparison of model behavior across different dataset groups.

The two dataset groups, corresponding to the 1:10k--1:25k and 1:10k--1:15k scale transitions, exhibit distinct impacts on model performance. In the 1:25k group, building geometries are more heavily generalized, resulting in predictions that closely approximate the overall structure of the ground truth but with reduced geometric detail. In contrast, the 1:15k group preserves finer architectural contours, which introduces increased geometric complexity into the prediction task.

Although predictions on the 1:15k group retain more detailed shapes, larger discrepancies with the ground truth are observed. This suggests that the models experience greater difficulty in handling datasets with higher geometric complexity, particularly in accurately modeling intricate outlines and maintaining consistent connectivity. These observations indicate that model performance is sensitive to the level of detail present in the input data, and that an intermediate degree of generalization may be more favorable for stable prediction outcomes.

\subsection{Impact Analysis of Features}
In addition to comparisons across different models and dataset groups, an ablation experiment was conducted to assess the influence of feature design on model performance. Two categories of features are considered: intrinsic geometric features and relative features describing spatial relationships.

To evaluate the contribution of relative features, model performance was compared under two configurations: with and without the inclusion of relative features. This comparison enables an assessment of how relative spatial information affects predictive accuracy in both link prediction and node movement tasks within the map generalization framework.

\begin{table}[h!]
\centering
\caption{Comparison of Validation Datasets Results for GraphSAGE}
\label{tab:validation_results}
{\footnotesize
\begin{tabularx}{\linewidth}{@{}lXX@{}}
\toprule
\textbf &\textbf{Link Prediction Accuracy} & \textbf{Node Movement MSE} \\
\midrule
With Relative Feature & \textbf{0.9559} & 2.1647 \\
Without Relative Feature & 0.9283 & \textbf{2.2009} \\
\bottomrule
\end{tabularx}
}
\end{table}

\begin{figure*}[h!]
    \centering
    \includegraphics[width=0.95\textwidth]{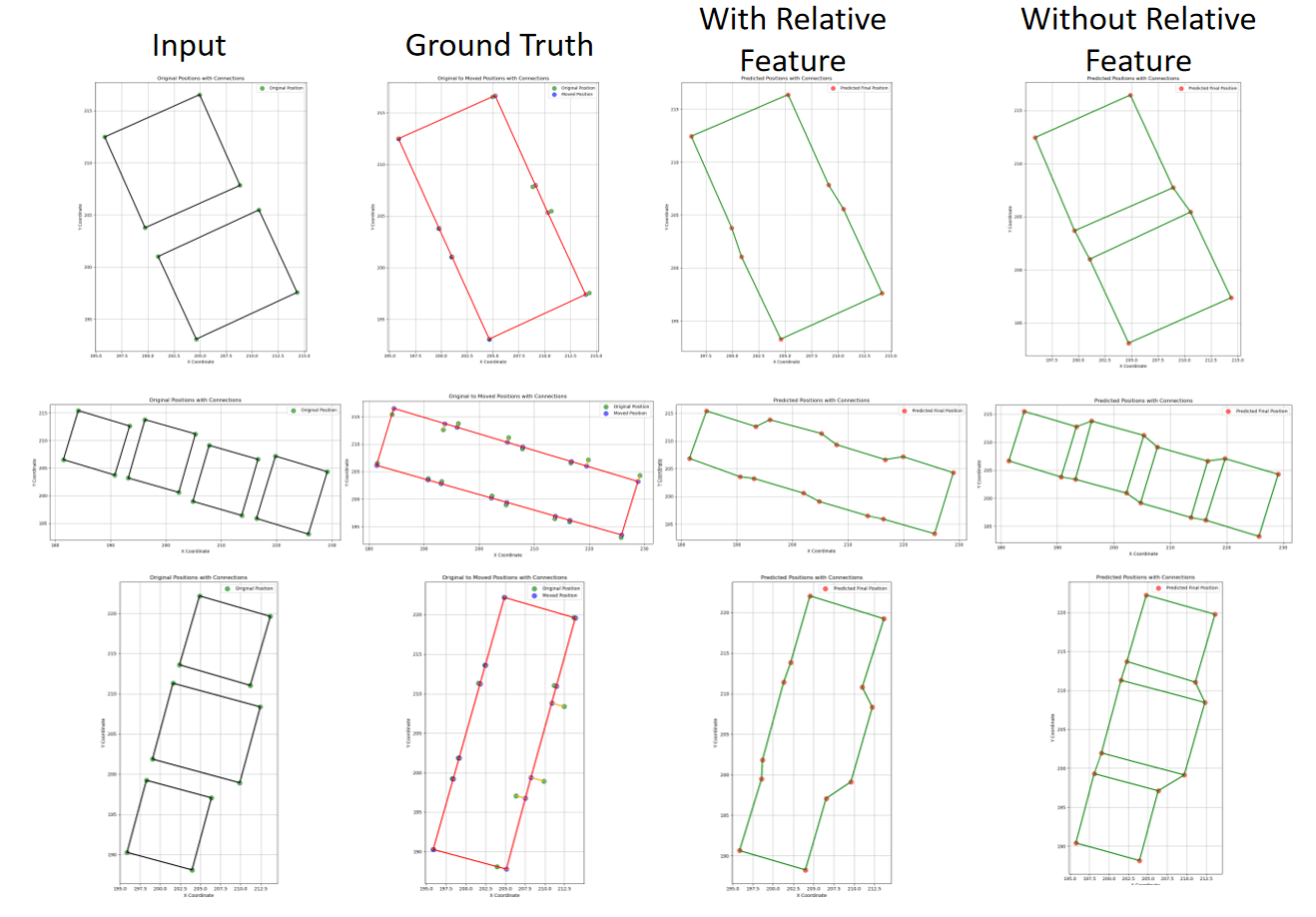}
    \caption{“with and without” Relative Feature Ablation Experiment Visualization Result}
    \label{FIG:18}
\end{figure*}

Figure~\ref{FIG:18} presents a qualitative comparison of prediction results obtained with and without the inclusion of relative features. Each row corresponds to an individual test case, showing the input data, the ground truth, and the predicted results under the two feature configurations. Predictions incorporating relative features generally exhibit closer alignment with the ground truth, particularly in terms of contour smoothness and edge removal. In contrast, predictions generated without relative features tend to display rougher outlines and larger deviations from the reference structures, indicating reduced geometric consistency.

The bar charts further illustrate node movement magnitudes in the horizontal ($\Delta x$) and vertical ($\Delta y$) directions across different test cases. Considerable variation in node displacement is observed, with a small number of nodes undergoing relatively large movements while most nodes exhibit minor changes. This uneven distribution suggests the presence of data imbalance in the node movement task, where models are more frequently exposed to small displacements during training.

Such imbalance may contribute to the limited predictive accuracy observed for nodes with larger displacements. In addition, the dataset includes one-to-many relationships arising from multiple generalization operators, including aggregation, simplification, selection, and typification. These heterogeneous transformation mechanisms cannot be fully distinguished based solely on spatial proximity, which further increases the difficulty of learning node movement patterns. Moreover, node displacement is modeled as an unconstrained regression problem, allowing predictions in arbitrary directions and magnitudes, which introduces additional challenges for stable learning. These observations indicate that more explicit constraints or task-specific regularization strategies may be beneficial for improving node movement prediction in future work.

\begin{figure}[h!]
	\centering
		\includegraphics[scale=.35]{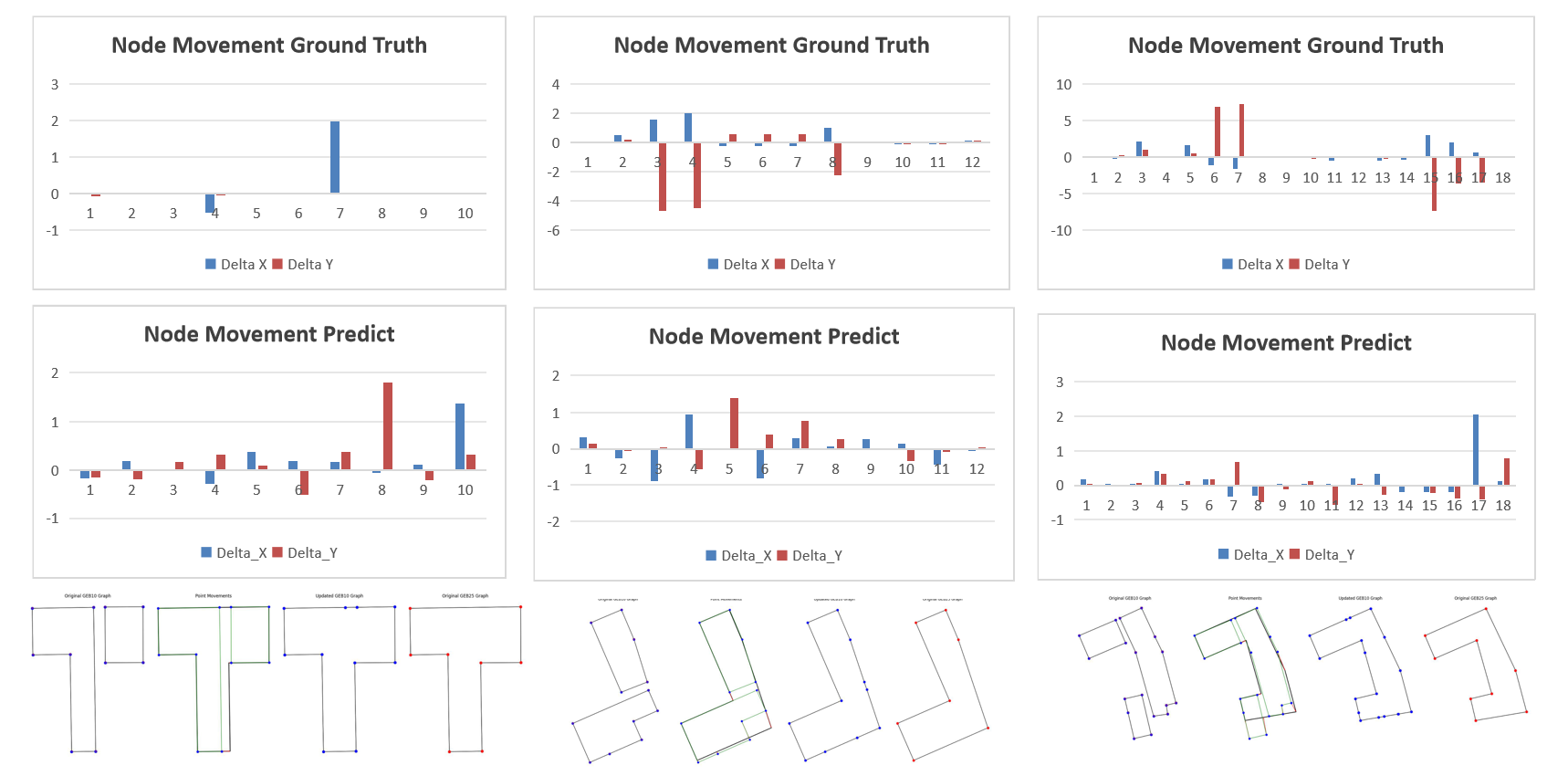}
	\caption{Data Imbalance Examples in node movement and link prediction task}
	\label{FIG:19}
\end{figure}

\section{Post Process and Evaluation}
\subsection{Post Process}

\begin{figure}[h!]
	\centering
		\includegraphics[scale=.4]{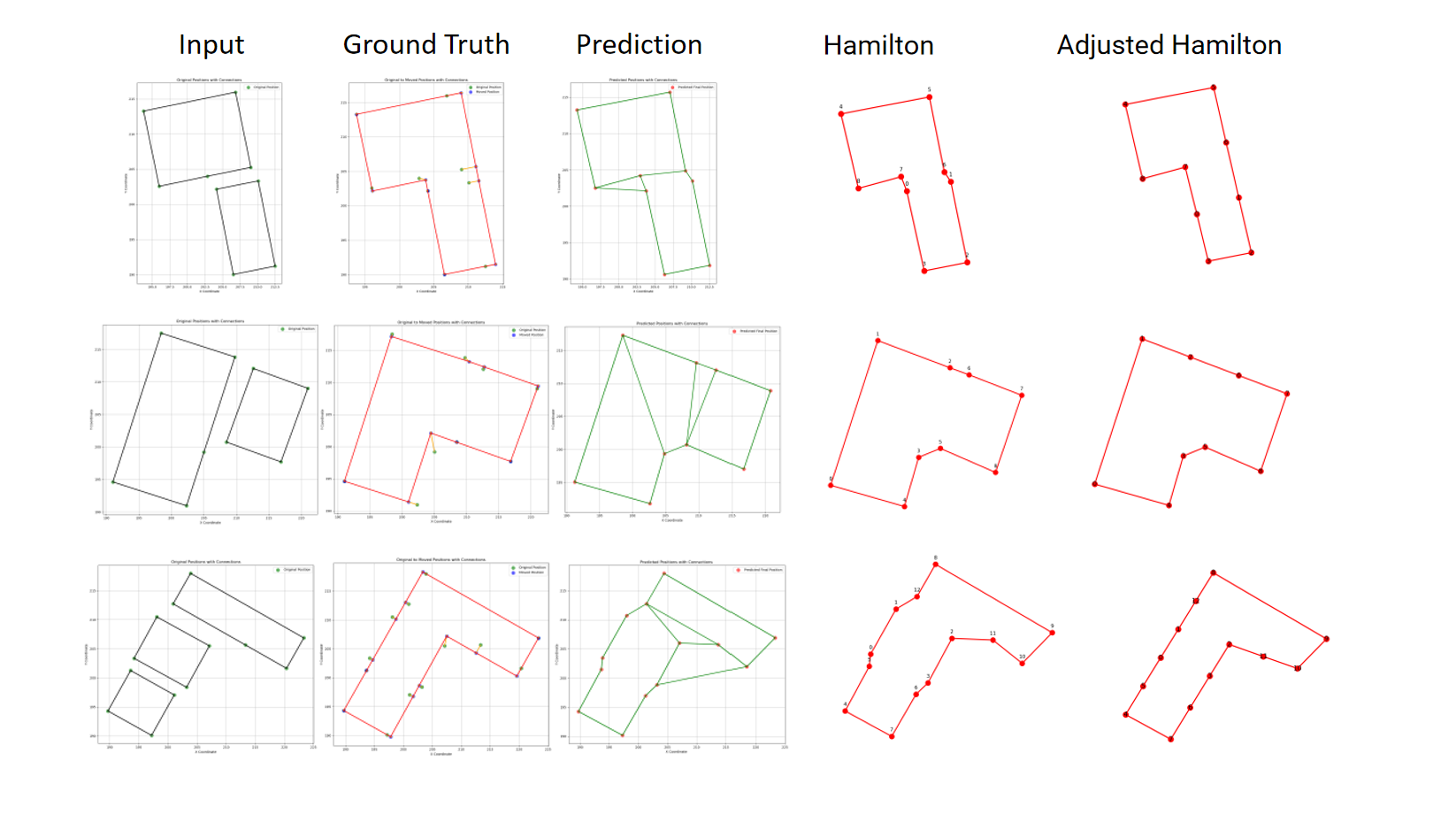}
	\caption{Hamiltonian cycle and angle-constrained Hamiltonian cycle used for post-processing}
	\label{fig:hamilton_cycle}
\end{figure}

While Hamiltonian cycle construction can enforce structural validity by removing redundant internal connections and restoring closed polygonal outlines, its applicability is inherently limited. In particular, predicted adjacency structures that contain missing essential connections or excessive structural inconsistencies cannot always be transformed into valid Hamiltonian cycles. Consequently, Hamiltonian cycle-based post-processing does not serve as a universal correction mechanism, but rather as a means of assessing whether predicted aggregation results are structurally recoverable.

Based on this consideration, the success rate of Hamiltonian cycle construction is reported as an evaluation metric. A prediction is regarded as successful if it either directly satisfies Hamiltonian cycle constraints or can be transformed into a single closed cycle through the removal of redundant connections. Predictions that fail to produce a valid cycle are classified as unsuccessful. This metric therefore reflects the degree to which predicted adjacency structures meet the topological requirements of valid building aggregation, rather than the raw accuracy of link prediction.

\begin{table}[htbp]
\centering
\caption{Success rate of aggregation results after Hamiltonian cycle-based post-processing}
\label{tab:hamilton_success_rate}
\begin{tabular}{lccc}
\toprule
\textbf{Dataset} & \textbf{GCN} & \textbf{GAT} & \textbf{GraphSAGE} \\
\midrule
1:15k Group & 43.99\% & 41.11\% & \textbf{55.77}\% \\
1:25k Group & 50.00\% & 51.60\% & \textbf{56.74}\% \\
\bottomrule
\end{tabular}
\end{table}

Table~\ref{tab:hamilton_success_rate} summarizes the Hamiltonian cycle success rates for GCN, GAT, and GraphSAGE across two dataset groups with different generalization levels. Across both scale groups, GraphSAGE yields higher success rates, indicating that a larger proportion of its predicted adjacency structures are structurally recoverable into valid aggregation results. GCN and GAT exhibit lower success rates, suggesting greater difficulty in producing topologically consistent outputs under the same post-processing criteria.

For all models, slightly higher success rates are observed for the 1:25k dataset compared to the 1:15k dataset. This trend suggests that stronger geometric generalization at coarser scales may reduce structural complexity and noise in predicted graphs, thereby increasing the likelihood that aggregation results satisfy Hamiltonian cycle constraints.

Overall, these results demonstrate that Hamiltonian cycle-based evaluation provides an effective means of assessing the structural soundness and recoverability of aggregation outputs. Rather than replacing learning-based predictions, this post-processing analysis complements quantitative link prediction metrics by explicitly evaluating whether predicted structures can be transformed into valid building outlines.

\subsection{Evaluation}

The test datasets are used to evaluate the performance of three graph neural network models—GCN, GAT, and GraphSAGE—on two tasks: link prediction and node movement.

For the link prediction task, model performance is evaluated using multiple metrics, including accuracy, F1 score, precision, recall, and closure ratio. As reported in Table~\ref{tab:link_prediction_metrics}, GraphSAGE achieves the highest accuracy and F1 score among the three models, while GAT yields comparable results with slightly lower values. GCN exhibits lower performance across most metrics.

GraphSAGE also attains higher precision and recall, indicating a more balanced ability to identify existing connections while limiting false positives. In addition, GraphSAGE achieves the highest closure ratio, suggesting that a larger proportion of its predicted adjacency structures form closed loops suitable for aggregation. GAT demonstrates similar but slightly lower closure performance, whereas GCN shows a comparatively reduced ability to generate closed structures.

Analysis of the confusion matrices further indicates that GraphSAGE and GAT produce higher counts of correctly predicted edges than GCN. Overall, these results suggest that GraphSAGE and GAT exhibit more stable and consistent performance in link prediction, with GraphSAGE showing a modest advantage across the evaluated metrics.

\begin{figure}[h!]
	\centering
		\includegraphics[scale=.4]{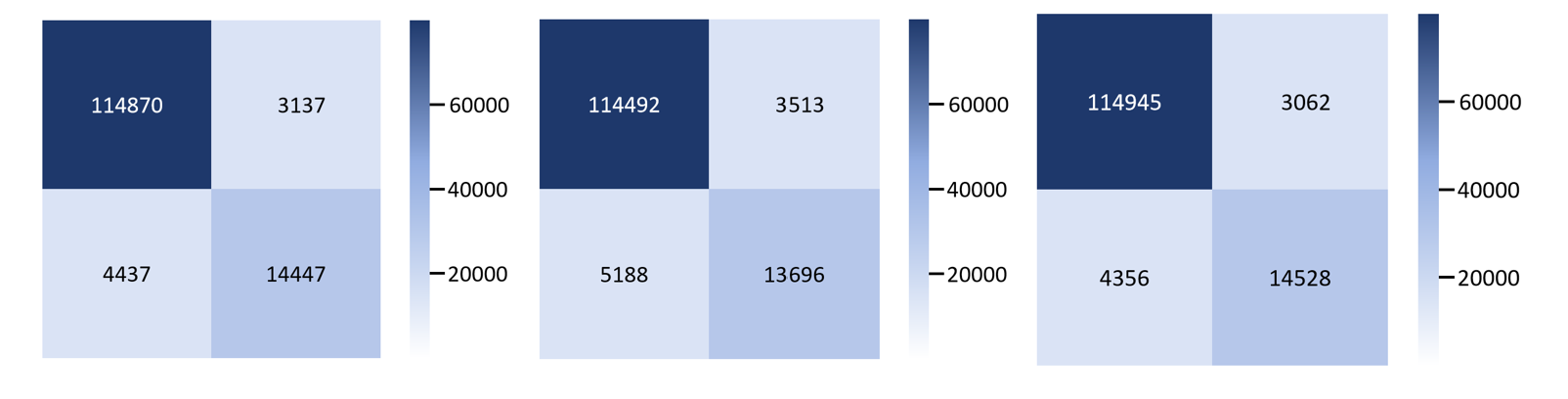}
	\caption{Visualization Confusion Matrix in Link Prediction Task with GAT, GCN, GraphSAGE Model }
	\label{FIG:1}
\end{figure}

\begin{table}[htbp]
\centering
\caption{Evaluation results of test datasets for the link prediction task}
\label{tab:link_prediction_metrics}
{\footnotesize
\begin{tabularx}{\linewidth}{@{}l*{5}{>{\centering\arraybackslash}X}@{}}
\toprule
\textbf{Model} & \textbf{Accuracy} & \textbf{F1 Score} & \textbf{Recall} & \textbf{Precision} & \textbf{Closure Ratio} \\
\midrule
GAT        & 0.9447 & 0.9438 & 0.9447 & 0.9433 & 0.8940 \\
GCN        & 0.9364 & 0.9352 & 0.9364 & 0.9344 & 0.8742 \\
GraphSAGE & \textbf{0.9458} & \textbf{0.9450} & \textbf{0.9458} & \textbf{0.9445} & \textbf{0.8958} \\
\bottomrule
\end{tabularx}
}
\end{table}

For the node movement task, model performance is evaluated using mean squared error (MSE), mean absolute error (MAE), and root mean squared error (RMSE). As reported in Table~\ref{tab:node_movement_metrics}, GraphSAGE achieves the lowest MSE among the three models, indicating smaller overall deviations from the ground truth compared to GAT and GCN. 

The MAE values are relatively close across models, with GAT exhibiting a slightly lower MAE than GraphSAGE and GCN, suggesting marginal differences in average absolute displacement errors. In terms of RMSE, GraphSAGE again attains the lowest value, reflecting reduced sensitivity to larger prediction errors. 

Overall, these results indicate that while average displacement errors are comparable across models, GraphSAGE demonstrates a more favorable error distribution, particularly in terms of squared-error-based metrics.

\begin{table}[htbp]
\centering
\caption{Evaluation results of test datasets for the node movement task}
\label{tab:node_movement_metrics}
{\footnotesize
\begin{tabularx}{\linewidth}{@{}l*{3}{>{\centering\arraybackslash}X}@{}}
\toprule
\textbf{Model} & \textbf{MSE} & \textbf{MAE} & \textbf{RMSE} \\
\midrule
GAT        & 1.3579 & 0.5465 & 1.1653 \\
GCN        & 1.3961 & 0.5487 & 1.1815 \\
GraphSAGE & \textbf{1.3276} & \textbf{0.5570} & \textbf{1.1522} \\
\bottomrule
\end{tabularx}
}
\end{table}

Across both link prediction and node movement tasks, GraphSAGE exhibits consistently strong performance across multiple evaluation metrics, including accuracy, precision, recall, and error-based measures. GAT demonstrates comparable performance, particularly in the link prediction task, while GCN shows relatively lower but still competitive results. These findings indicate clear performance differences among the evaluated graph neural network architectures, with GraphSAGE and GAT exhibiting more stable and robust behavior across tasks.

\begin{figure}[h!]
	\centering
		\includegraphics[scale=.7]{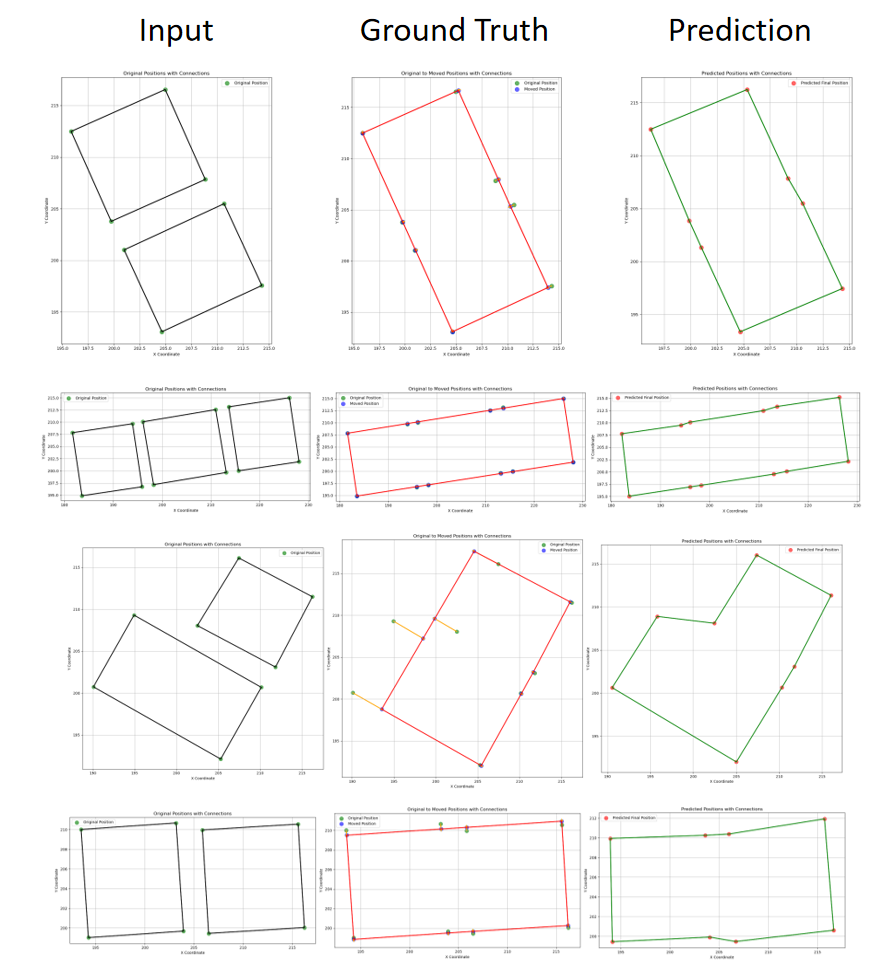}
	\caption{Visualization Results of the GraphSAGE Model on the Test Datasets}
	\label{FIG:22}
\end{figure}

Figure~\ref{FIG:22} presents representative qualitative examples of predictions generated by the GraphSAGE model on the test dataset. The visualizations illustrate the model’s ability to capture overall aggregation structures through link prediction, as well as its behavior in estimating node movement under varying geometric configurations. These examples provide qualitative evidence that complements the quantitative evaluation results.

\section{Conclusion and Discussion}

This study explores the feasibility of applying graph-based deep learning methods to automated map generalization, with a focus on the simplification and aggregation of vector-based building footprints. By reformulating simplification as a node movement task and aggregation as a link prediction task, the problem is embedded within a unified graph learning framework, enabling a systematic comparison of representative graph neural network architectures.

The experimental results indicate that graph neural networks are capable of capturing certain geometric and structural patterns relevant to map generalization. However, the findings also reveal several limitations that constrain predictive performance. In particular, model behavior is sensitive to feature design, and the ability to accurately predict large node displacements and complex topological changes remains limited. Ablation experiments further demonstrate that relative and contextual features describing spatial relationships can improve performance in both node movement and link prediction tasks, although these improvements are not consistent across models or datasets. This variability suggests that effective feature selection and update mechanisms remain a nontrivial challenge.

The comparison between single-task and multi-task training highlights additional trade-offs between modeling efficiency and predictive accuracy. While joint training provides a unified learning framework, it does not consistently outperform task-specific training, indicating potential interference between regression-based node movement and classification-based link prediction objectives. These results suggest that task coupling in graph-based generalization requires careful design to avoid performance degradation.

Post-processing using traditional algorithms, including Hamiltonian cycle construction, convex hulls, and genetic algorithms, further illustrates the potential of hybrid approaches that combine learning-based predictions with algorithmic structural constraints. Although such methods can improve the structural validity of predicted aggregation results, their reliance on post hoc correction and manual judgment underscores current limitations in achieving fully automated generalization pipelines.

Overall, the results suggest that graph deep learning provides a feasible but not yet sufficient pathway for automatic map generalization. Challenges such as data imbalance, limited representation of rare events, and difficulties in preserving topological consistency under complex transformations continue to affect model performance. Rather than offering a definitive solution, this study contributes methodological insights into how simplification and aggregation can be formulated as graph learning problems, and highlights the strengths and limitations of current graph neural network approaches.

Future work may explore more expressive feature update mechanisms, data augmentation or reweighting strategies to address imbalance, and tighter integration of learning-based models with domain-specific rules and constraints. By systematically analyzing both quantitative performance and structural validity, this study provides a foundation for further research toward more robust and interpretable graph-based map generalization systems.



\bibliographystyle{cas-model2-names}
\bibliography{main}


\end{document}